\documentclass{article} %

\usepackage[final]{colm2026_conference}

\usepackage[T1]{fontenc}
\usepackage{tabularx,makecell} %
\usepackage[english,bidi=default]{babel}
\babelprovide[import]{yiddish}

\DeclareFontFamily{TU}{freeyiddish}{}

\DeclareFontShape{TU}{freeyiddish}{m}{n}
  {<-> "[FreeSerif.otf]:script=hebr"}{}

\DeclareFontShape{TU}{freeyiddish}{b}{n}
  {<-> "[FreeSerifBold.otf]:script=hebr"}{}

\DeclareFontShape{TU}{freeyiddish}{m}{it}
  {<-> "[FreeSerifItalic.otf]:script=hebr"}{}

\DeclareFontShape{TU}{freeyiddish}{b}{it}
  {<-> "[FreeSerifBoldItalic.otf]:script=hebr"}{}

\DeclareFontShape{TU}{freeyiddish}{bx}{n}
  {<-> ssub * freeyiddish/b/n}{}

\DeclareFontShape{TU}{freeyiddish}{bx}{it}
  {<-> ssub * freeyiddish/b/it}{}

\newcommand{\texthebrew}[1]{%
  \foreignlanguage{yiddish}{%
    \fontencoding{TU}%
    \fontfamily{freeyiddish}%
    \selectfont
    #1%
  }%
}

\usepackage{microtype}
\usepackage{hyperref}
\usepackage{url}
\usepackage{booktabs}
\usepackage{multirow}
\usepackage{xcolor}
\usepackage{float}
\usepackage{amsmath}
\usepackage{tikz}
\usepackage{emo}
\usepackage{bxcoloremoji}
\usepackage{wrapfig}

\usetikzlibrary{positioning,arrows.meta,calc}

\newcommand{\corpus}{\textit{Oytser}} %
\newcommand{\eval}{\textit{Kashes}} %
\newcommand{\mllm}{\textsc{MameLoshnLM}}

\usepackage{lineno}
\usepackage{todonotes}
\usepackage{multirow}
\usepackage{enumitem}

\usepackage{graphicx}
\usepackage{subcaption}
\usepackage{caption}
\usepackage{fontawesome5}

\definecolor{darkblue}{rgb}{0, 0, 0.5}
\hypersetup{colorlinks=true, citecolor=darkblue, linkcolor=darkblue, urlcolor=darkblue}

\definecolor{forestgreen}{rgb}{0.13, 0.55, 0.13}

\let\svthefootnote\thefootnote
\newcommand\freefootnote[1]{%
  \let\thefootnote\relax%
  \footnotetext{#1}%
  \let\thefootnote\svthefootnote%
}

\title{\mllm: Yiddish Language Model \\ and Evaluation Benchmark}

\makeatletter
\renewcommand{\@fnsymbol}[1]{%
  \ifcase#1\or \text{ }\else\@arabic{#1}\fi}
\makeatother

\author{
Uri Katz$^{1}$\thanks{Correspondence to \texttt{urikacid@gmail.com}.  \text{\texthebrew{אָ}} Equal senior authors.} \;
Omer Goldman$^{2}$ \;
Tomasz Limisiewicz$^{3}$ \\
\textbf{
Reut Tsarfaty$^{1,\text{\texthebrew{אָ}}}$ \
Noah A. Smith$^{3,4,\text{\texthebrew{אָ}}}$} \\
$^{1}$Bar-Ilan University \quad
$^{2}$University of Cambridge \\
$^{3}$University of Washington \quad
$^{4}$Allen Institute for AI \\
}

\begin{document}
\ifcolmsubmission
\linenumbers
\fi

\maketitle

\begin{abstract}
We present \mllm,\footnote{%
  \begin{tabular}[t]{@{}l@{}}
    \faIcon{github}\enspace
    \url{https://github.com/katzurik/MameLoshnLM} \\[1pt]
    \coloremojicode{:hugging:}\enspace
    \url{https://huggingface.co/Yiddish-NLP}
  \end{tabular}%
}
the first open-source 8B-parameter language model built specifically for Yiddish. Despite Yiddish's rich textual tradition, its limited digital presence and the scarcity of reliable evaluation resources have constrained progress in Yiddish language modeling. Existing multilingual corpora and benchmarks are often poor proxies for the language, containing substantial amounts of noisy, machine-translated, and misclassified text. We address these gaps by introducing \corpus, a high-quality Yiddish pretraining corpus that combines contemporary web-native sources with literary materials, and \eval, a multi-task benchmark spanning translation, linguistic analysis, information extraction, and language understanding. Using these resources, we continue pretraining Llama 3.1 8B to obtain \mllm. Across the tasks in the benchmark, \mllm\ outperforms open baselines of similar scale.
Our analyses show that these gains are not only quantitative: relative to general-purpose multilingual models, \mllm\ better captures language-defining lexical and morphological patterns, pointing to a broader failure mode of noisy web-scale multilingual data for low-resource languages. Our results provide both a foundation for Yiddish NLP and a practical template for language model development in historically rich but digitally underrepresented languages.

\end{abstract}

\section{Introduction}

Despite swift advancements in multilingual language modeling, large language models' (LLMs)
benefits remain highly uneven across languages \citep{wu2025bitterlessonlearned2000}. Performance is typically strongest for languages with extensive digital presence, large quantities of high-quality text, and mature evaluation resources, while low-resource languages remain substantially underserved. Yiddish is a particularly interesting case. Historically a Germanic language written primarily in Hebrew script, it also includes a substantial Hebrew and Aramaic lexical layer and reflects long-standing contact with Slavic languages. Although it has a long and rich literary history, its contemporary online presence is limited, and much of the Yiddish text available in common web-scale datasets is sparse, noisy, or poorly matched to the language as it is 
used today. The result is models that serve neither the language's roughly one million speakers nor the academic community that researches the history and culture of the Yiddish-speaking world.

These limitations reflect gaps across the full LM development pipeline. On the data side, the amount of publicly available Yiddish pretraining text is limited \citep{xue-etal-2021-mt5}, and its quality has not been well characterized \citep{kreutzer-etal-2022-quality}. In this paper, we show through a detailed analysis of mC4 that much of its putative Yiddish content is either machine-translated spam or not Yiddish at all.
We find that less than half of the data is genuine Yiddish text. On the evaluation side, benchmarks for Yiddish are scarce, and those that do exist are often based on automatic translation rather than tasks designed for the language itself \citep{singh-etal-2024-aya}. Addressing Yiddish effectively therefore requires more than simply adding more text: it requires coordinated work on corpus construction, evaluation, model adaptation, and analysis.

We introduce \mllm, an open-source 8B-parameter LM for Yiddish, together with two supporting resources: \corpus, a new Yiddish pretraining corpus, and \eval, a benchmark for evaluating Yiddish LMs. \corpus\ was designed as a higher-quality alternative to common open multilingual resources by combining contemporary web-native Yiddish with materials drawn from Yiddish's literary tradition. This is particularly important for Yiddish, whose historical textual record is far richer than its current web footprint. We also introduce \eval, a benchmark comprising existing and newly developed tasks, intended to provide a broader and more reliable basis for evaluating Yiddish language models.

Using these resources, we continue pretraining Llama~3.1~8B to obtain \mllm. Across the benchmark, \mllm\ outperforms open models of similar scale, with strong gains on Yiddish-centered tasks such as translation, linguistic analysis, and named entity recognition. Beyond benchmark performance, our analysis shows that \mllm\ outputs more natural Yiddish, morphologically and lexically, avoiding overly Germanized and translationese-like characteristics often produced by general-purpose multilingual LMs (\autoref{sec:analysis}). We further find that, while English data is important for preserving the broader capabilities of the base model during continued pretraining, adding data from historically or genealogically related languages yields only limited and inconsistent gains (\autoref{subsec:mixtures}).

More broadly, our results point to a challenge that extends beyond Yiddish. For historically rich but digitally underrepresented languages, the main obstacle to language-model development may be not only data scarcity, but mismatch between authentic language use and the noisy public web data on which multilingual models are often trained. We present Yiddish as a clear case of this problem, and \mllm, \corpus, and \eval\ as a practical demonstration that targeted resource construction and continued pretraining can substantially narrow the gap.

\section{Related Work}
Recent work has shown that continued pretraining of strong open-weight models is an effective approach for adapting LMs to low- to moderate-resource languages, including, for example, Basque, Estonian, Kazakh, and Setswana \citep{etxaniz-etal-2024-latxa,kuulmets-etal-2024-teaching,koto2025sherkalachat, brown-marivate-2025-pula}. However, this paradigm still leaves open important questions about training data composition \citep{zhang-etal-2025-snakmodel}, the role of language mixing \citep{elhady-etal-2025-emergent}, and the value of machine-translated versus native text in low-resource adaptation \citep{doshi-etal-2024-pretraining}. Our work follows this general approach for the low-resource language Yiddish, but examines it in a particularly constrained setting, marked by limited training data, low-quality publicly available text, and very limited evaluation resources.

Although Yiddish has received only limited attention in NLP, several important efforts have created the foundations on which modern work can build. Early work addressed phrase-based machine translation \citep{genzel-etal-2009-creating}, while later efforts developed foundational infrastructure such as a Yiddish speech corpus \citep{webber2022reyd,bleaman2025corpus}, and a basic ASR system  \citep{cavar-etal-2016-generating}. Other work supported core NLP building blocks through resources and models for part-of-speech tagging \citep{santorini2021penn, kulick2022part}, orthographic variation, and transliteration \citep{saleva-2020-multi}. More recently, Jochre 3 \citep{urieli2025jochre} substantially improved OCR for printed Yiddish, enabling the digitization of thousands of Yiddish books and supporting the computational use of large historical collections such as the Yiddish Book Center digital library. In addition, some Yiddish resources now exist within broader multilingual efforts, such as dependency annotation in UD \citep{yiddishUD} and domain-specific NER in EHRI \citep{dermentzi-scheithauer-2024-repurposing}, but no prior work, to our knowledge, has targeted Yiddish through the lens of modern LM development. Our work aims to fill this gap by combining a dedicated large-scale pretraining corpus, a broad benchmark for Yiddish language models, and an open-source Yiddish model.
\section{Training Data}

The quality of pretraining data is a central challenge for low resource LM development \citep{kreutzer-etal-2022-quality, doshi-etal-2024-pretraining, ali-etal-2025-judging}. This section first examines Yiddish in existing open corpora, focusing on an analysis of the Yiddish portion of mC4 to quantify the extent of noise, machine-translated content, and misclassified non-Yiddish text. It then introduces \corpus, our new Yiddish pretraining corpus, which was designed to provide a substantially cleaner and more representative alternative.

\subsection{Yiddish in Existing Corpora}
\label{sec:mc4-analysis}

We investigate the amount and quality of Yiddish texts in open source corpora.
For that purpose, we focus on mC4 \citep{xue-etal-2021-mt5}, one of the most prominent open multilingual pretraining corpora. mC4 was introduced as the training corpus for mT5 and is based on filtered Common Crawl data covering 101 languages. After filtering, the corpus totals 6.6B pages and 6.3T tokens. 

Within mC4, approximately 0.3B words and 0.1M pages are tagged as Yiddish. This makes mC4 a potentially attractive default source of Yiddish data for multilingual pretraining. However, these aggregate statistics provide no information about the linguistic quality of the split itself.

We therefore carried out an analysis of the Yiddish portion of mC4.\footnote{We ran our analysis on the AI2 implementation of mC4, available at \url{https://huggingface.co/datasets/allenai/c4}.} To estimate the amount of machine-translated material, we inspected source URLs in the Yiddish split and identified pages containing Yiddish ISO codes that appeared to be one language-specific rendering among many automatically generated versions of the same site. We complemented this with a corpus-wide pass of an additional language identifier, and with manual assessment of hundreds of domains and sampled pages. Full technical details of the audit are given in \autoref{app:mc4-audit}.

Overall, our analysis suggests that fewer than half (42.2\%) of the documents in the Yiddish mC4 split are genuine high-quality Yiddish from validated native sources. At least 29.8\% of the documents appear to be machine-translated, and an additional 21.9\% are Hebrew texts mistakenly identified as Yiddish, and the remaining 6\% is a long tail of short fragments, multilingual pages,
and small uncatalogued Yiddish sources.
\begin{wraptable}{r}{0.43\textwidth}
  \small
  \setlength{\tabcolsep}{3pt}
  \begin{tabular}{@{}lrr@{}}
    \toprule
    Category & Pages & \% of split \\
    \midrule
    Known Yiddish sources  & 60,587 & 42.2\% \\
    Machine-translated     & 42,765 & 29.8\% \\
    Misidentified Hebrew   & 31,485 & 21.9\% \\
    Other                  & 8,871  &  6.2\% \\
    \bottomrule
  \end{tabular}
  \caption{Yiddish in mC4.}
  \label{tab:mc4-composition}
\end{wraptable}
These findings show that the existence of a Yiddish split in a large multilingual corpus should not be equated with the availability of native Yiddish pretraining data. Instead of relying on large automatically constructed Common Crawl corpora, we chose to build \corpus\ as a genuinely high-quality and verified alternative, described in the next subsection.

\begin{table}[t]
\centering
\small
\begin{tabular}{llrrr}
\toprule
Source & Type & Documents & Words & Tokens \\
\midrule
YBC & books/literature & 12.3K & 723.09M & 4.23B \\
Ivelt & discussion forum & 47.9K & 148.49M & 797.20M \\
Kaveshtiebel & discussion forum & 12.3K & 28.49M & 154.53M \\
Forward & news/magazine & 10.3K & 6.72M & 49.41M \\
Wikipedia & encyclopedia & 15.3K & 2.77M & 16.87M \\
Hamaspik & community newspaper & 78 & 2.24M & 13.88M \\
yiddish.news & news & 12.0K & 2.10M & 13.51M \\
Lebns Fragn & news/media & 1.42K & 0.76M & 5.53M \\
The Hebrew Bible & religious text & 37 & 0.55M & 3.67M \\
\midrule
\textbf{Total} & --- & \textbf{346.3K} & \textbf{915.21M} & \textbf{5.28B} \\
\bottomrule
\end{tabular}
\caption{Statistics of \corpus, the Yiddish training corpus by source and type. The token count is according to \mllm's tokenizer.}
\label{tab:corpus_sources}
\end{table}

\subsection{Data Sources}
\label{subsec:corpus}

We constructed \corpus\ (\texthebrew{אוצר}, ``treasure'') a new Yiddish pretraining corpus designed as a higher-quality alternative to existing open resources. Our goal was to assemble a corpus that  reflects genuine Yiddish usage across domains and registers. The corpus combines contemporary casual texts with literary materials, reflecting the online footprint of present-day Yiddish and its much richer textual tradition. Our corpus is constructed from two source groups, each serving a distinct role:

\paragraph{\textbf{Web-native Yiddish sources.}} This portion of the data is composed of a wide range of contemporary and diverse forms of Yiddish. These include Yiddish news websites and magazines, the Yiddish Wikipedia, and a Yiddish translation of the Hebrew Bible; in addition, we included several Yiddish forums, where users write in a freer style that is less well-conformed with the standard. In general, the data includes more than 0.3M native Yiddish documents from various topics, dialects, and communities.

\paragraph{\textbf{Yiddish Book Center Corpus (YBC)}} Our largest data source is the Yiddish Digital Library of the YBC,\footnote{\url{https://www.yiddishbookcenter.org/collections/digital-yiddish-library}} whose books were OCRed using Jochre3 \citep{urieli2025jochre}. With more than 12K books and approximately 720M Yiddish words, this is, to our knowledge, the largest and most comprehensive digital source of Yiddish texts currently available. This source allows us to draw on the long literary history of Yiddish, rather than relying only on its limited contemporary web presence. Most of the books in this collection date to the last 100 years, making it an important source for capturing the stylistic, dialectal, generic, and historical breadth of Yiddish, spanning not only literary texts but also essays, historical works, and other prose genres.

Together, these resources provide coverage of multiple genres and registers. \autoref{tab:corpus_sources} summarizes the data sources included in our corpus, together with their genres and sizes. All in all, the corpus contains more than 915M words from 9 high quality online and literary sources.

\paragraph{\textbf{Licensing and collection.}} The data used for continued pretraining combines licensed, public-domain, and publicly accessible Yiddish-language sources. The digitized books were used under our research agreement with the Yiddish Book Center, Yiddish Wikipedia under the applicable Wikimedia licenses, Lebns Fragn approved for non-commercial research, the Forward archive was used for non-commercial research consistent with the fair-use provision in its terms of use, and the Yehoash Yiddish translation of the Hebrew Bible from a CC0 public-domain source. The remaining web-native data was collected using source-specific extraction pipelines from domains already represented in Common Crawl. Collection was limited to pages accessible without login, subscription, or paywall and followed the applicable site-level crawling directives at the time of extraction. Further legal and ethical considerations are discussed in the Ethics Statement.

\paragraph{\textbf{De-identification of user-generated data.}} Before training, we applied a corpus-wide de-identification pipeline using carefully designed regular-expression patterns, consistent with the rule-based PII filtering approaches used in ROOTS and Dolma \citep{bigscience-roots:2022,soldaini-etal-2024-dolma}. These filters were used to remove email addresses, phone numbers, URLs, user handles, and non-date numerical sequences that could encode identifiers.

For the Ivelt and Kaveshtiebel forums, we additionally retained only post and comment text and removed all account and interaction metadata, including author usernames, timestamps, signatures, thread identifiers, and reply relations. The text was then separated from its original discussion context, segmented, and shuffled, so that user profiles, account histories, and conversation graphs were not preserved. We iteratively sampled and inspected the processed forum data, refining the filters when recurring patterns were identified. In the final inspected samples, we found no explicit personal information that could readily be linked to a specific user or forum account.

\section{\eval: The Yiddish Evaluation Benchmark}
For low-resource languages like Yiddish, the scarcity of evaluation data is as limiting as the scarcity of training data.
Available Yiddish evaluation resources are few and scattered, and to the best of our knowledge, no prior work has systematically benchmarked language models across multiple Yiddish tasks.

We introduce \eval\ (\texthebrew{קאַשעס}, ``difficult questions''), the first multi-task evaluation benchmark for Yiddish language models.
The tasks in \eval\ reflect both the applicative needs of Yiddish scholars and digital humanities researchers, such as linguistic analysis, information extraction, as well as general language model capabilities like translation, commonsense reasoning and paraphrase detection.
In addition to consolidating existing resources, we contribute a new parallel corpus for machine translation -- the largest natively authored Yiddish--English benchmark to date.

\begin{table}[t]
\centering
\small
\begin{tabular}{llrl}
\toprule
\textbf{Task} & \textbf{Dataset} & \textbf{Size} & \textbf{Source}  \\
\midrule
\multirow{2}{*}{Translation} & \eval-mt & 5{,}287 & \textcolor{purple}{[new task]} see \autoref{par:kashes-mt}   \\
 & FLORES+ & 1{,}012 & \citet{nllb-24}  \\
\midrule
POS tagging & \multirow{4}{*}{Yiddish Treebank} & 1{,}079 & \multirow{4}{*}{\citet{yiddishUD}}  \\
Dep.\ parsing &  & 1{,}079 &   \\
Transliteration &  & 1{,}079 &   \\
Lemmatization &  & 955 &  \\
\midrule
\multirow{3}{*}{NER} & EHRI-NER & 4{,}103 & \citet{dermentzi-scheithauer-2024-repurposing}  \\
 & WikiANN & 300 & \citet{rahimi-etal-2019-massively}  \\
 & newNLP NER & 1{,}535 & \citet{berkovitch_rusinek_yiddish_2022}  \\
\midrule
Commonsense QA & PIQA & 625 & \multirow{3}{*}{ \citet{singh-etal-2024-aya}} \\
Question generation & WikiQA & 293 &   \\
Paraphrase det. & PAWS-Wiki & 8000 &   \\
\bottomrule
\end{tabular}
\caption{Datasets included in \eval. Size denotes the number of samples (sentences or sentence pairs).}
\label{tab:eval-tasks}
\end{table}

\autoref{tab:eval-tasks} summarizes all datasets included in \eval.
The benchmark spans 9 tasks:
Four linguistic analysis tasks (POS tagging, lemmatization, dependency parsing, and transliteration) are drawn from the YiTB treebank \citep{yiddishUD}, with structured annotations flattened into sequence-level formats for generative models.
Named entity recognition is evaluated on three datasets: EHRI-NER \citep{dermentzi-scheithauer-2024-repurposing} with annotated Holocaust-related documents, WikiANN \citep{rahimi-etal-2019-massively} with tagged entities from the 2018 Yiddish Wikipedia dump, and newNLP NER \citep{berkovitch_rusinek_yiddish_2022} with texts from historical Yiddish newspapers annotated by scholars. 

Three language-understanding tasks are taken from the Aya machine-translated collection \citep{singh-etal-2024-aya}: PIQA, a two-choice commonsense reasoning task; WikiQA, question generation from a Wikipedia paragraph; and PAWS-Wiki, paraphrase detection. Because there are currently no native Yiddish benchmarks for general language understanding, we use these translated tasks to cover evaluation types otherwise unavailable in Yiddish, retaining only those Aya tasks whose translations preserved the essential structure and intent of the original.

\subsection{\eval-mt: New Yiddish Translation Benchmark}
\label{par:kashes-mt}

Existing Yiddish--English parallel corpora, such as FLORES+ \citep{nllb-24}, are derived from English source texts translated into Yiddish.
Therefore, they do not capture the authentic voice of native Yiddish writing.
We address this gap by creating two sentence-level corpora from bilingual online publications, where the Yiddish texts are originally written by native speakers and the English translations are produced by Yiddish scholars and professional translators.

We collect human-translated documents from two web sources:
(1). \emph{Forverts}\footnote{\url{https://forward.com/yiddish/}} is a digital newspaper containing articles and blog posts from the past decade for which corresponding English versions are available. Both language versions are authored by native speakers. 
(2). \emph{In geveb}\footnote{\url{https://ingeveb.org/}} is a peer-reviewed, open-access
journal of Yiddish studies that publishes Yiddish literature alongside English
translations produced by Yiddish scholars and professional literary translators.

\begin{figure}[t]
\centering
\resizebox{\textwidth}{!}{%
\begin{tikzpicture}[
  font=\small,
  node distance=6mm and 9mm,
  source/.style={draw, rounded corners=2pt, fill=blue!8, minimum width=27mm, minimum height=10mm, align=center},
  stage/.style={draw, rounded corners=2pt, fill=gray!10, minimum width=21mm, minimum height=11mm, align=center},
  final/.style={draw, thick, rounded corners=2pt, fill=blue!15, minimum width=22mm, minimum height=11mm, align=center},
  side/.style={draw, dashed, rounded corners=2pt, fill=orange!10, minimum width=32mm, align=center, font=\footnotesize},
  counts/.style={font=\scriptsize, text=black!60, align=center},
  pairlab/.style={font=\scriptsize\itshape, text=black!70, align=center, fill=white, inner sep=1pt},
  arr/.style={-{Latex[length=2mm]}, thick}
]
 
\node[source] (yid) {\textbf{Yiddish documents}\\\scriptsize natively authored};
\node[source, below=11mm of yid] (eng) {\textbf{English versions}\\\scriptsize by scholars \& translators};
 
\draw[{Latex[length=1.6mm]}-{Latex[length=1.6mm]}, dashed, black!60]
  (yid.south) -- node[pairlab]{same articles,\\two editions} (eng.north);
 
\node[counts, below=1.5mm of eng] {\textit{Forverts, In geveb}};
 
\node[stage, right=13mm of $(yid.east)!0.5!(eng.east)$] (docmatch)
  {Document\\matching\\\scriptsize (TF-IDF over titles\\\scriptsize / shared doc IDs)};
 
\node[stage, right=of docmatch] (align)
  {Sentence\\alignment\\\scriptsize (SentAlign:\\\scriptsize LaBSE similarity)};
\node[stage, right=of align] (docfilt)
  {Document\\filtering\\\scriptsize ($\geq$35\% aligned)};
\node[stage, right=of docfilt] (sentfilt)
  {Sentence\\filtering\\\scriptsize (score $\geq 0.65$)};
\node[stage, right=of sentfilt] (dedup) {De-\\duplication};
 
\node[final, right=of dedup] (out) {\textbf{Kashes-mt}\\\scriptsize 5,287 pairs};
 
\draw[arr] (yid.east) -- (docmatch.west|-yid.east) -- (docmatch);
\draw[arr] (eng.east) -- (docmatch.west|-eng.east) -- (docmatch);
 
\draw[arr] (docmatch) -- (align);
\draw[arr] (align) -- (docfilt);
\draw[arr] (docfilt) -- (sentfilt);
\draw[arr] (sentfilt) -- (dedup);
\draw[arr] (dedup) -- (out);
 
\node[counts, below=2mm of align] (c1) {6,952};
\node[counts] at (docfilt |- c1) {6,176};
\node[counts] at (sentfilt |- c1) {5,358};
\node[counts] at (dedup |- c1) {5,287};
\node[counts, left=3mm of c1] {\textit{aligned sentences:}};
 
\node[side, below=11mm of out.south west, anchor=north east, xshift=6mm] (leak)
  {source documents removed\\from \textit{Oytser} pretraining corpus};
\draw[arr, dashed] (out.south) |- (leak.east);
 
\end{tikzpicture}

}
\caption{The \eval-mt construction pipeline. Yiddish-English editions
of the same articles are matched and sentence-aligned to construct a
natively authored translation benchmark}
\label{fig:kashes-pipeline}
\end{figure}
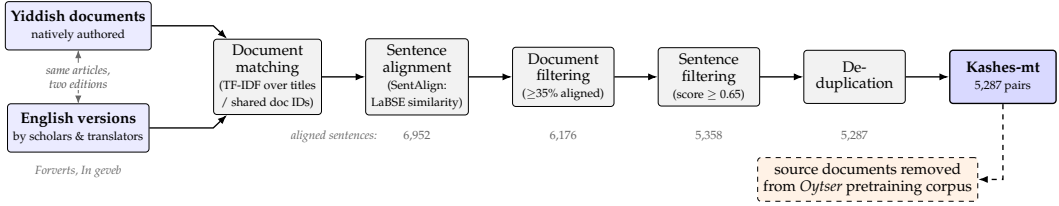

We construct \eval-mt using a four-stage pipeline:
\textbf{document matching}, \textbf{sentence alignment},
\textbf{quality filtering}, and \textbf{deduplication}
(\autoref{fig:kashes-pipeline}). For \textit{In geveb}, bilingual editions
share document IDs; for \textit{Forverts}, we match similar but nonidentical
English titles using TF-IDF. SentAlign \citep{sentalign-2023} then aligns
sentences within matched documents. We discard documents with fewer than
$35\%$ aligned sentences and sentence pairs with a SentAlign score below
$0.65$,\footnote{Score distributions are reported in
\autoref{sec:sentalign-scores}.} and remove duplicate pairs. The resulting
benchmark contains 995 pairs from \textit{Forverts} and 4{,}292 from
\textit{In geveb}, for 5{,}287 high-quality pairs in total; 
\autoref{fig:kashes-pipeline} reports retention at each stage. To prevent
benchmark leakage, we exclude every source document from the \corpus{}
pretraining corpus \citep{jacovi-etal-2023-stop,
balloccu-etal-2024-leak}.

\section{\mllm}
We present \mllm, the first open-source large language model for Yiddish, trained on the Yiddish corpus described in \autoref{subsec:corpus}. We name the model \emph{Mame-Loshn} (\texthebrew{מאַמע־לשון}, ``mother tongue'') in reference to the traditional Yiddish term that evokes the language's intimate connection to home and family. We show that it outperforms strong baselines of similar scale across a broad set of Yiddish evaluation benchmarks.

\paragraph{Training Details.}

\mllm\ was produced by continued pretraining of Llama-3.1-8B on a Yiddish corpus using a causal language modeling objective. The model was trained in bfloat16 precision using the 8-bit AdamW optimizer with a learning rate of $2 \times 10^{-5}$, a cosine learning rate scheduler with a 2\% warmup ratio, and a weight decay of 0.01. Training was conducted for one epoch with a maximum sequence length of 1,024 tokens, a per-device batch size of 38, and 4 gradient accumulation steps, yielding an effective batch size of approximately 155K tokens. To mitigate catastrophic forgetting of the base model's broader capabilities, we include English data from CC100 \citep{conneau-etal-2020-unsupervised} during continued pretraining in a Yiddish-dominant mixture: Yiddish accounts for 72\% of words (92\% of tokens), and English for the remaining 28\% (8\% of tokens). We also tested multilingual variants that reallocate part of this non-Yiddish budget to historically related languages, namely German, Hebrew, Polish and Russian (See \autoref{subsec:mixtures}). In total, the model was trained on approximately 5.7 billion tokens over 36,663 optimization steps. Gradient checkpointing was enabled to reduce memory consumption. Training was conducted on a single NVIDIA H200 GPU for approximately 207 GPU-hours.

\section{Experimental Setup}

We evaluate \mllm{} alongside five open-weight baseline models of similar scale: Llama~3.1~8B \citep{grattafiori2024llama}, Qwen3~8B \citep{yang2025qwen3}, BLOOMZ~7B \citep{muennighoff2022crosslingual}, Gemma-2~9B \citep{gemmateam2024gemma2improvingopen}, and EuroLLM~9B \citep{martins2024eurollmmultilinguallanguagemodels}. Qwen3 is the only baseline whose documentation explicitly lists Yiddish as a supported language. The remaining models do not claim explicit Yiddish support, but provide strong, comparable-size open-weight baselines with varying degrees of multilingual capability; Llama~3.1 additionally provides the direct comparison to \mllm{}'s base model. To our knowledge, none has previously been systematically evaluated across a broad Yiddish benchmark. We evaluate all models under identical 1-, 3-, and 5-shot conditions, sampling demonstrations with the same random seed in a leave-one-out scheme. Example inputs and demonstrations are provided in \autoref{app:kashes-examples}.

\section{Results}
\subsection{Performance on \eval}
We report 5-shot results across all tasks and models in \autoref{tab:main-results}, with full results for all shot settings (1, 3, and 5-shot) in \autoref{app:full-results}.

Across 14 evaluations, \mllm{} achieves the best overall performance, with the highest average score (62.6), ahead of Gemma-2 9B (57.0), Llama 3.1 8B (56.8), and Qwen3 8B (54.7). Its gains are concentrated on Yiddish-centered tasks: it leads on POS tagging, dependency parsing, transliteration, EHRI and newNLP NER.

\begin{table}[t]
\centering
\scriptsize
\setlength{\tabcolsep}{4pt}
\begin{tabular}{llcccccc}
\toprule
Dataset & Metric & \mllm & Llama 3.1 8B & Qwen3 8B & BLOOMZ 7B & Gemma-2 9B & EuroLLM 9B \\
\midrule

POS Tagging & Accuracy & \textbf{88.6} & 86.9 & 85.9 & 12.9 & 87.6 & 68.2 \\
Dep.\ Parsing & LAS & \textbf{40.6} & 39.7 & 40.3 & 3.8 & 36.9 & 14.8 \\
Lemmatization & Change Acc. & 31.9 & 19.7 & 21.6 & 3.4 & \textbf{36.5} & 11.2 \\
Transliteration & 1--CER & \textbf{92.3} & 92.1 & 88.5 & 20.5 & 91.6 & 90.5 \\
PAWS-Wiki & Accuracy & 62.9 & 55.8 & \textbf{75.2} & 43.0 & 61.4 & 54.6 \\
PIQA & Accuracy & 47.1 & 45.0 & \textbf{50.3} & 16.5 & 14.2 & 46.8 \\
Wiki QA & ROUGE-L & 34.4 & 32.3 & \textbf{34.8} & 16.8 & 34.2 & 27.8 \\

\midrule
\multicolumn{8}{c}{\textit{Named Entity Recognition}}\\[-1pt]
\midrule
EHRI & Micro F1 & \textbf{41.3} & 34.2 & 20.8 & 0.8 & 31.7 & 9.2 \\
WikiANN & Micro F1 & 59.7 & 58.1 & 54.6 & 13.2 & \textbf{62.8} & 40.8 \\
newNLP & Micro F1 & \textbf{57.6} & 51.9 & 50.5 & 6.4 & 50.9 & 35.8 \\

\midrule
\multicolumn{8}{c}{\textit{Machine Translation: English $\to$ Yiddish}} \\[-1pt]
\midrule
FLORES+ & COMET & \textbf{78.5} & 64.8 & 46.5 & 32.9 & 66.7 & 47.2 \\
\eval-mt & COMET & \textbf{75.3} & 59.6 & 45.3 & 33.9 & 59.9 & 45.5 \\[-1pt]

\midrule
\multicolumn{8}{c}{\textit{Machine Translation: Yiddish $\to$ English}} \\[-1pt]
\midrule
FLORES+ & COMET & \textbf{87.2} & 82.2 & 79.1 & 44.8 & 86.1 & 75.6 \\
\eval-mt & COMET & \textbf{79.5} & 72.8 & 72.0 & 44.4 & 77.3 & 70.7 \\

\midrule
\textbf{Average} & & \textbf{62.6} & 56.8 & 54.7 & 20.9 & 57.0 & 45.6 \\
\bottomrule
\end{tabular}
\caption{Evaluation results on \eval\ benchmark. \textbf{Bold} = best model per row. All metrics: higher is better.}
\label{tab:main-results}
\end{table}

It also remains competitive on the remaining tasks, including near-best performance on WikiANN NER and WikiQA. The main exceptions are PAWS-Wiki and PIQA, which are Aya benchmarks based on machine-translated versions of widely used multilingual datasets.
The performance of \mllm{} is especially strong in the English to Yiddish translation (more than 11 COMET points improvement over the closest competitor). 
It indicates that continued training enabled the generation of fluent texts in Yiddish. %
We further support this claim by qualitative analysis of sentences translated into Yiddish in \autoref{sec:analysis}.

Overall, the results indicate that continuing pretraining in authentic Yiddish data yields substantial improvements in tasks that require lexical, orthographic, and syntactic command of Yiddish, while maintaining competitive performance on more general benchmarks.

\subsection{Effect of related-language mixing.}
\label{subsec:mixtures}
A practical question in low-resource continued pretraining is how to use the non-target-language portion of the training budget \citep[cf.][]{koto2025sherkalachat}. For Yiddish, a natural hypothesis is that historically related languages such as Hebrew and German may be more useful than English, either because they share lexical material with Yiddish or because they may support transfer on linguistically relevant structures. We test this directly by keeping the same Yiddish corpus and continued-pretraining setup as \mllm, and training two variants that modify only the remaining mixture. Data for each language was drawn from CC100 \citep{conneau-etal-2020-unsupervised} in fixed-size blocks. The two configurations were designed to test whether reallocating part of the English budget to related languages (\textit{Rebalanced}) or supplementing it with additional multilingual data (\textit{Expanded}) would improve Yiddish adaptation. In \textit{Rebalanced}, part of the English budget is redistributed to Hebrew and German; in \textit{Expanded}, we add Hebrew, German, Russian, and Polish while keeping the absolute English amount (exact mixture proportions in \autoref{app:mixtures}). Because Yiddish tokenizes much more densely than the other languages, these word-level changes reduce the Yiddish share of training tokens from 91.8\% in \mllm\ to 85.2\% and 71.0\%, respectively. Although the multilingual variants remain competitive in translation and show isolated gains, they do not yield a consistently better Yiddish model, and on several Yiddish-centered tasks they underperform not only \mllm\ but even the base Llama 3.1 model (See \autoref{app:mixtures} for full results). These findings suggest that, in our setting, maintaining a strongly Yiddish-dominant training signal with a limited amount of English is a more effective strategy than reallocating that budget to related languages.

\section{Analysis}
\label{sec:analysis}

Yiddish is a useful test case for analyzing what multilingual language models learn in low-resource settings. A general multilingual model can often produce something understandable in Yiddish because Yiddish is typically represented, at least to some extent, in the training data, and also overlaps with neighboring languages and shares script with Hebrew. But this surface fluency can be misleading: models may produce plausible Yiddish while still missing many of the features that make the language sound native. This makes Yiddish a useful case for a broader question: when multilingual models appear to handle a low-resource language, are they learning the language itself, or a flatter approximation shaped by noisy data?

\begin{table}[t]
\centering
\small
\begin{tabular}{lcccc}
\toprule
Probe & \mllm & Llama 3.1 8B & Gold & $p$ \\
\midrule
\multicolumn{5}{l}{\textit{LK vocabulary (Eng$\rightarrow$Yid translation)}} \\
LK content word rate (\%) & 4.7 & 1.6 & 6.2 & $< 10^{-161}$ \\
LK sentence match rate (\%) & \textbf{52.4} & 16.0 & --- & $< 10^{-229}$ \\
\midrule
\multicolumn{5}{l}{\textit{Morphology (lemmatization change accuracy)}} \\
\textit{ge-} participles ($n$=236) & \textbf{50.8} & 5.1 & --- & $< 10^{-29}$ \\
Hebrew-origin plurals ($n$=53) & \textbf{30.2} & 2.3 & --- & $< 0.001$ \\
Determiner paradigm ($n$=291) & \textbf{28.2} & 15.5 & --- & $< 10^{-5}$ \\
\midrule
\multicolumn{5}{l}{\textit{Auxiliary control pair}} \\
\texthebrew{זײַן} ``to be'' --- suppletive ($n$=460) & \textbf{20.0} & 8.9 & --- & $< 10^{-9}$ \\
\texthebrew{האָבן} ``to have'' --- regular ($n$=163) & 84.7 & 84.7 & --- & 1.0 (n.s.) \\
\bottomrule
\end{tabular}
\caption{Linguistic competence probes, \mllm\ vs.\ Llama 3.1 8B (5-shot).
Translation metrics are computed over \eval-mt sentence pairs;
morphological metrics are lemmatization change accuracy on the UD Yiddish-YiTB
test set. Gold denotes the rate in native references. Full results in
\autoref{tab:full-summary}.}
\label{tab:analysis-summary}
\end{table}

We examine this question directly by probing phenomena that distinguish native Yiddish from such an approximation. Our analysis\footnote{For reproducibility, implementation details for all analysis steps are provided in \autoref{app:analysis-methodology}.} focuses on two cases: the lexical layer that is derived from Hebrew/Aramaic (known as \emph{loshn-koydesh}, abbreviated hereafter LK) and Yiddish-specific morphology. Across both, general multilingual models often produce an intelligible but systematically reduced variety of Yiddish, while \mllm\ more closely matches patterns found in native Yiddish data.  

\paragraph{General multilingual models underproduce core \emph{loshn-koydesh} vocabulary.}
A defining property of Yiddish is its substantial Hebrew lexical layer, which includes many frequent everyday words. If a model has learned native-like Yiddish, it should produce these items naturally. If instead it has only shallow or noisy knowledge of Yiddish, these words should be systematically underproduced, avoided, or replaced. We test this directly in English$\rightarrow$Yiddish translation of the \eval-mt benchmark using the loshn-koydesh pronunciation lexicon\footnote{Based on Niborski's Lexicon of Loshn-Koydesh words, \url{https://github.com/ibleaman/loshn-koydesh-pronunciation}.} to identify LK words. In 5,287 5-shot translations, gold references contain LK words in 6.2\% of content tokens. \mllm\ produces LK words at 4.7\%, whereas Llama~3.1~8B produces only 1.6\% ($p < 10^{-229}$, paired t-test). The same pattern appears at the sentence level: when a gold translation contains an LK word, \mllm\ is much more likely than Llama to produce a matching LK item in its translation of the same input (52.4\% vs.\ 16.0\% in 5-shot). Importantly, this effect is not simply a reflection of overall translation quality. Per-sentence LK recall correlates only weakly with COMET score (Spearman $\rho = 0.23$,$p < 0.001$), suggesting that COMET is largely blind to loss of the \emph{loshn-koydesh} lexical layer. A translation can therefore score well on overall quality while replacing Hebrew-origin vocabulary with Germanic alternatives. This shows that Llama misses many LK words that appear in native Yiddish renderings of the same content. Moreover, Llama's LK output is concentrated in proper nouns and culturally salient items such as \texthebrew{ישׂראל} (Israel), \texthebrew{רבי} (Rebbe), and \texthebrew{תּורה} (Torah), rather than in the common vocabulary that characterizes natural Yiddish usage. See Appendix \autoref{tab:lk-top15} for the most frequent LK words and their frequencies in both models.

This deficit is not random. When the reference contains an LK noun with a Germanic alternative, Llama often substitutes the Germanic form, e.g., \texthebrew{משפּחה} ``family'' with \texthebrew{פֿאַמיליע}, \texthebrew{מלחמה} ``war'' with \texthebrew{קריג}, and \texthebrew{פּנים} ``face'' with \texthebrew{געזיכט}. For frequent LK function words with no simple Germanic equivalent, such as \texthebrew{אפֿשר} ``perhaps,'' \texthebrew{כּמעט} ``almost,'' and \texthebrew{כּדי} ``in order to,'' Llama often avoids the lexical item through paraphrase or restructuring. The result is often understandable, but it is noticeably more Germanized and less native-like than either the reference or the output of \mllm\ . The contrast therefore concerns not only how much Yiddish the models produce, but what kind of Yiddish they produce.

\paragraph{The same non-native pattern appears in Yiddish-specific morphology.}
If general multilingual models have only limited command of Yiddish-specific morphology, they should also fail on irregular inflectional patterns, where the correct lemma cannot be recovered by simple surface copying. This is exactly what we observe in lemmatization. The strongest contrast appears on past participles, normally formed with a \textit{ge-} prefix but rife with irregularities. On 236 such tokens, \mllm\ reaches 50.8\% accuracy, while Llama reaches only 5.1\% ($p < 10^{-29}$). Similar gaps appear on Hebrew-origin plural patterns such as ``tales'' \texthebrew{מעשׂיות}$\rightarrow$\texthebrew{מעשׂה} (30.2\% vs.\ 2.3\%), and on the definite article paradigm, where surface forms such as \texthebrew{די}, \texthebrew{דאָס}, and \texthebrew{דעם}, all corresponding to English \textit{the}, must be mapped to the same citation lemma \texthebrew{דער} (28.2\% vs.\ 15.5\%).

The same contrast appears within auxiliary verbs. On regular forms of \texthebrew{האָבן}, such as \texthebrew{האָט}$\rightarrow$\texthebrew{האָבן}, the two models perform identically (84.7\% vs.\ 84.7\%). This suggests that both can handle cases where the lemma remains locally recoverable from the surface form. But on forms of the highly irregular verb \texthebrew{זײַן} ``to be,'' whose inflected forms often look very different from the lemma, \mllm\ performs much better than Llama (20.0\% vs.\ 8.9\%). In these cases, Llama's dominant strategy is to copy the input unchanged, suggesting that it often fails to recover the underlying paradigm. The gap is therefore not uniform across morphology: it is largest when successful lemmatization requires knowledge of lexeme-specific inflectional system rather than simple regular transformations.

\paragraph{This pattern is consistent with the data-quality picture in \autoref{sec:mc4-analysis}.}
As discussed above, publicly available Yiddish web corpora contain substantial amounts of machine-translated material, weakening the already limited signal available for learning native-like Yiddish. Our mC4 audit provides a direct measurement of this effect: machine-translated pages in the Yiddish split show an LK rate of 3.6\%, less than half the 10.2\% of validated native sources (\autoref{app:mc4-audit}).
Llama's own LK production rate falls below even that of the machine-translated web text. In generation, the model reproduces the same lexical depletion that characterizes its likely training data.
General multilingual models can thus produce superficially fluent Yiddish while missing the lexical and morphological features that make the language sound native. Continued pretraining on curated Yiddish text reduces this gap and recovers much of the missing competence. More broadly, this suggests a problem that may extend to other low-resource languages with limited web presence: multilingual web corpora can support surface-level fluency while still weakening language-specific properties that matter for authentic generation.

\section{Discussion and Conclusion}

A natural question is why a language like Yiddish needs a dedicated language model. Yiddish is spoken by roughly one million people worldwide, and its textual heritage is vast and increasingly the subject of active research. Yet without adequate language technologies, this heritage remains difficult to search, organize, and analyze at scale. Notably, several datasets in \eval, such as EHRI-NER and newNLP NER, were created by digital humanities  scholars whose primary goal is extracting information from Holocaust testimonies and historical newspapers, not evaluating language models. Improved performance on these tasks therefore reflects not just benchmark gains but progress toward tools that these communities can actually use.

Beyond its practical importance, Yiddish is an informative case for multilingual NLP. Its position at the intersection of several language families, combined with its unique script, makes it a compelling testbed for research on tokenization, cross-lingual transfer, multilingual data mixing, and the effects of data scarcity and quality on language model development. This problem likely extends to other languages whose authentic usage diverges from their web footprint, including languages in diglossia situations and those with fragmented online presence. 

This work presents the first comprehensive LLM development effort for Yiddish, encompassing corpus construction, benchmark curation, model training, and evaluation. Building on this foundation, future work can pursue instruction tuning for interactive use, the creation of additional training and evaluation resources, and the application of \mllm\ to large-scale digital humanities workflows. We hope that the work presented here, including \mllm, \corpus, and \eval, will contribute to the Yiddish-speaking and research communities and provide a useful reference point for similar efforts in other underrepresented languages.

\section*{Acknowledgments}

The authors would like to thank the Yiddish Book Center for its permission to use the Steven Spielberg Digital Yiddish Library, and \emph{In geveb: A Journal of Yiddish Studies} for granting permission to use its expert translation materials as part of our evaluation benchmark.
OG research is funded by the Blavatnik Family Foundation, his work on this paper was also partially funded by UniDive COST Action (\#CA21167).

\section*{Ethics Statement}
\paragraph{Purpose and institutional partnerships.}
Yiddish is a language of significant historical and cultural importance that remains severely underrepresented in modern NLP resources. This work aims to support Yiddish-language preservation, accessibility, and research, and was developed in partnership with institutions dedicated to Yiddish culture. The Yiddish Book Center licensed the Steven Spielberg Digital Yiddish Library for model training, and \textit{In geveb} licensed material used in our evaluation benchmark. MameLoshnLM is released under a non-commercial license consistent with the terms of these agreements.
\paragraph{Copyright and fair-use considerations.} The corpus combines material used under institutional agreements, open and public-domain licenses, applicable source terms, and publicly accessible web text. More than 82\% of the training tokens derive from sources covered by the first three categories. The web-native material was collected for non-commercial academic research exclusively from pages accessible without login, subscription, paywall, or circumvention of technical access controls, and in accordance with applicable site-level crawling directives. We verified that these domains were already represented in Common Crawl and in widely used multilingual corpora derived from it, including mC4 and OSCAR \citep{xue-etal-2021-mt5,abadji-etal-2022-towards}. Our source-specific pipelines therefore provide a cleaner and more complete extraction of Yiddish text from established public-web sources rather than exposing previously inaccessible material.

Consistent with the considerations applied in Dolma \citep{soldaini-etal-2024-dolma} to training on publicly available web data, we rely on fair use and analogous research exceptions as the legal basis for this training-stage use, which is non-expressive and transformative: a computational use intended to learn general linguistic patterns rather than to provide access to or substitute for individual works. This position is supported by the treatment of non-commercial research and analysis under United States fair-use principles and by the Israeli Ministry of Justice's conclusion that machine-learning uses will generally fall within fair-use and related statutory exceptions \citep{usco2025ai,israelMOJ2022}.

\paragraph{Personal information and privacy.}
To our knowledge, Ivelt and Kaveshtiebel are the only substantial publicly accessible digital sources we identified for contemporary, informal, community-authored Yiddish. Their inclusion was important for representing everyday language that is largely absent from historical and edited collections. They are also already prominent in Common Crawl-derived training data: in the Yiddish portion of mC4, Ivelt is the most frequent source domain and Kaveshtiebel is the fifth most frequent.
Following the harm-minimization approach recommended by the Association of Internet Researchers \citep{franzke2020aoir}, we applied source-specific safeguards beyond the corpus-wide processing described in Section~3.2. We removed account and interaction structure, retained only decontextualized post and comment text, and iteratively inspected processed samples for identifying information, refining the procedure whenever recurring risks were found. No explicit personal information readily linkable to a specific user or account was found in the final inspected samples. These measures were intended to preserve the linguistic value of the data while minimizing disclosure and re-identification risks.

\bibliography{colm2026_conference,anthology-1,anthology-2}
\bibliographystyle{colm2026_conference}

\appendix
\section{\eval-mt Sentence Alignment}
\label{sec:sentalign-scores}

\begin{figure}[!htb]
\begin{subfigure}{0.45\textwidth}
    \centering
    \includegraphics[width=\linewidth]{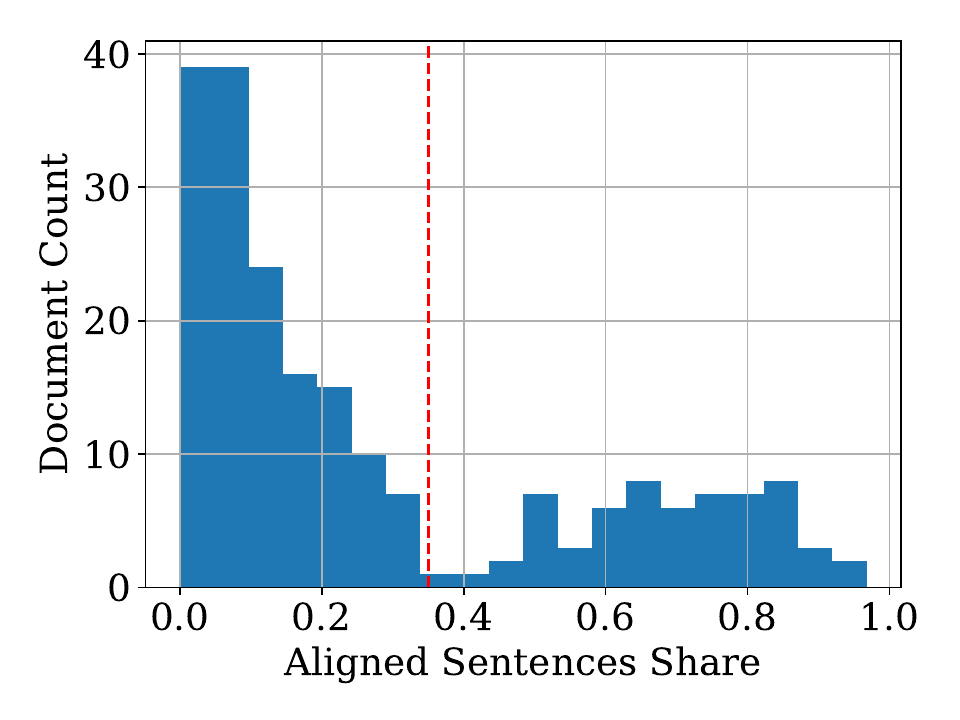}
    \caption{Forverts}
\end{subfigure}
\hfill
\begin{subfigure}{0.45\textwidth}
    \centering
    \includegraphics[width=\linewidth]{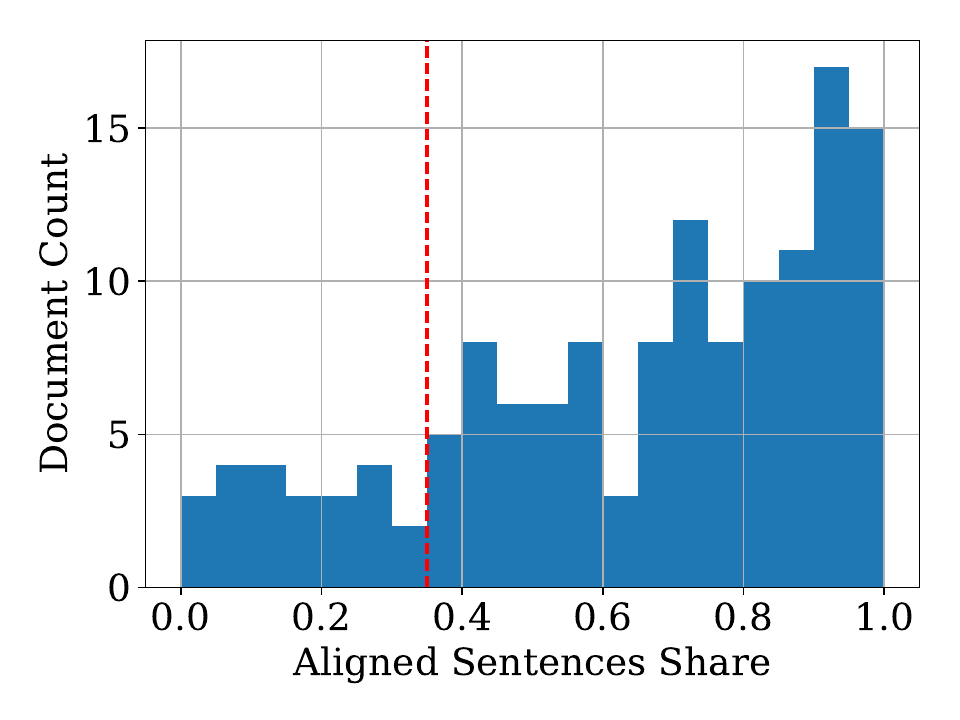}
    \caption{In geveb}
\end{subfigure}
\caption{The distribution of the portion of initially aligned sentences in all sentences in each document. We selected documents with at least $35\%$ sentences aligned to filter out documents that could be incorrectly matched.}
\label{fig:sentaling-doc}
\end{figure}

\begin{figure}[!htb]
\begin{subfigure}{0.45\textwidth}
    \centering
    \includegraphics[width=\linewidth]{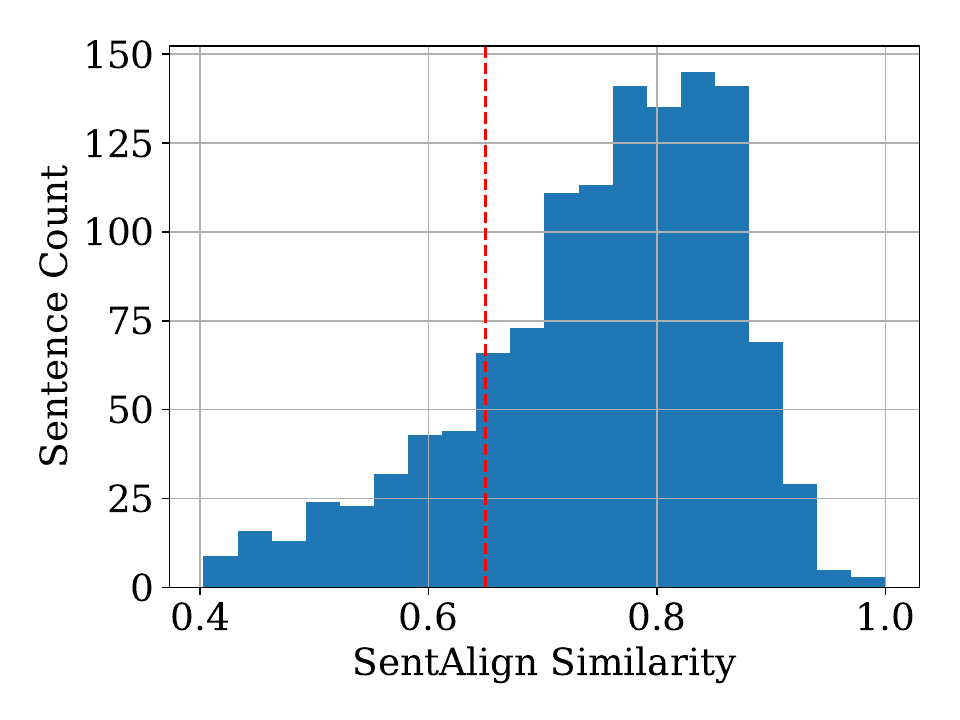}
    \caption{Forverts}
\end{subfigure}
\hfill
\begin{subfigure}{0.45\textwidth}
    \centering
    \includegraphics[width=\linewidth]{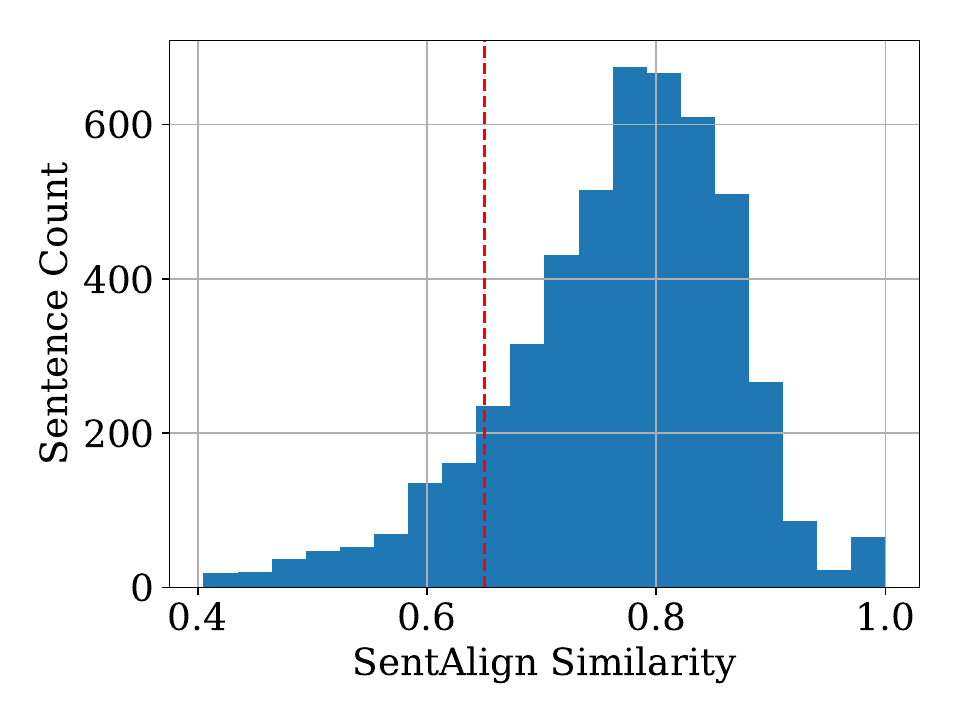}
    \caption{In geveb}
\end{subfigure}
\caption{The distribution of SentAlign similarity score across initially aligned documents. We selected a threshold of 0.65 to filter out pairs that could be an inaccurate translation.}
\label{fig:sentaling-scores}
\end{figure}

\autoref{fig:sentaling-doc} shows the distribution of the share of sentences that were initially aligned across all documents.
We filter out documents with less than $35\%$ sentences aligned.
In \autoref{fig:sentaling-scores}, we present the distribution of similarity scores between pairs of sentences aligned with SentAlign \citep{sentalign-2023}. 
We filter sentences with scores below $0.65$ that make up the lower tail of the distribution.

\section{Few-shot full result}
\label{app:full-results}
\begin{table}[H]
\centering
\caption{Results across all tasks and shot counts. \textbf{Bold} = best per row.}
\tiny
\setlength{\tabcolsep}{3pt}
\begin{tabular}{llcccccc}
\toprule
Task & Metric & \mllm\ & Llama 3.1 8B & Qwen3 8B & BLOOMZ 7B & Gemma-2 9B & EuroLLM 9B \\
\midrule
\multicolumn{8}{c}{\textit{1-shot}} \\
\midrule
POS Tagging & Accuracy & 83.3 & 83.9 & 84.3 & 6.8 & \textbf{88.3} & 67.7 \\
Dep. Parsing & LAS & 14.8 & 28.4 & \textbf{35.4} & 3.9 & 35.4 & 7.0 \\
Lemmatization & Change Acc. & 7.3 & 7.8 & 12.7 & 0.2 & \textbf{27.9} & 1.5 \\
Transliteration & 1-CER & \textbf{91.6} & 90.5 & 87.1 & 19.9 & 90.0 & 90.6 \\
PAWS-Wiki & Accuracy & \textbf{55.8} & \textbf{55.8} & 51.1 & 55.7 & 55.6 & 49.3 \\
PIQA & Accuracy & 46.3 & 26.9 & \textbf{53.7} & 11.8 & 20.7 & 20.7 \\
Wiki QA & ROUGE-L & 23.9 & 21.8 & \textbf{31.2} & 11.0 & 23.1 & 17.5 \\
NER (EHRI) & Micro F1 & 15.8 & 15.5 & 16.5 & 1.2 & \textbf{17.2} & 12.2 \\
NER (WikiANN) & Micro F1 & 27.5 & \textbf{31.5} & 30.7 & 4.8 & 28.5 & 18.9 \\
NER (newNLP) & Micro F1 & 19.2 & 20.3 & \textbf{28.5} & 2.8 & 25.0 & 19.0 \\
MT Eng$\to$Yid (FLORES+) & COMET & \textbf{77.9} & 63.0 & 43.0 & 33.0 & 64.2 & 45.0 \\
MT Eng$\to$Yid (\eval-mt) & COMET & \textbf{73.7} & 57.3 & 43.1 & 36.7 & 57.6 & 44.2 \\
MT Yid$\to$Eng (FLORES+) & COMET & \textbf{86.8} & 80.5 & 76.8 & 43.9 & 85.1 & 73.4 \\
MT Yid$\to$Eng (\eval-mt) & COMET & \textbf{78.8} & 70.7 & 70.3 & 43.0 & 76.0 & 68.5 \\
\midrule
\textit{Average} & & \textbf{50.2} & 46.7 & 47.5 & 19.6 & 49.6 & 38.2 \\
\midrule
\multicolumn{8}{c}{\textit{3-shot}} \\
\midrule
POS Tagging & Accuracy & 86.9 & 86.5 & 85.9 & 7.9 & \textbf{89.1} & 70.1 \\
Dep. Parsing & LAS & 36.9 & 37.5 & \textbf{38.2} & 2.5 & 36.3 & 11.1 \\
Lemmatization & Change Acc. & 18.3 & 12.3 & 10.5 & 0.6 & \textbf{22.8} & 4.5 \\
Transliteration & 1-CER & 92.0 & \textbf{92.0} & 87.9 & 21.7 & 91.5 & 89.8 \\
PAWS-Wiki & Accuracy & 59.8 & 55.8 & \textbf{75.1} & 49.8 & 61.4 & 47.7 \\
PIQA & Accuracy & \textbf{47.2} & 41.6 & 14.0 & 15.9 & 14.7 & 46.0 \\
Wiki QA & ROUGE-L & 33.0 & 30.2 & 33.8 & 15.7 & \textbf{34.0} & 26.5 \\
NER (EHRI) & Micro F1 & \textbf{50.0} & 37.8 & 17.6 & 1.1 & 27.9 & 16.9 \\
NER (WikiANN) & Micro F1 & \textbf{56.5} & 53.2 & 47.7 & 7.8 & 55.3 & 31.6 \\
NER (newNLP) & Micro F1 & 46.9 & 48.0 & \textbf{49.9} & 5.6 & 46.2 & 33.6 \\
MT Eng$\to$Yid (FLORES+) & COMET & \textbf{78.2} & 64.3 & 45.5 & 32.4 & 65.9 & 46.3 \\
MT Eng$\to$Yid (\eval-mt) & COMET & \textbf{75.0} & 58.9 & 44.6 & 34.3 & 59.2 & 45.0 \\
MT Yid$\to$Eng (FLORES+) & COMET & \textbf{87.1} & 81.8 & 78.8 & 44.7 & 85.8 & 75.2 \\
MT Yid$\to$Eng (\eval-mt) & COMET & \textbf{79.3} & 72.2 & 71.6 & 44.0 & 77.0 & 70.1 \\
\midrule
\textit{Average} & & \textbf{60.5} & 55.2 & 50.1 & 20.3 & 54.8 & 43.9 \\
\midrule
\multicolumn{8}{c}{\textit{5-shot}} \\
\midrule
POS Tagging & Accuracy & \textbf{88.6} & 86.9 & 85.9 & 12.9 & 87.6 & 68.2 \\
Dep. Parsing & LAS & \textbf{40.6} & 39.7 & 40.3 & 3.8 & 36.9 & 14.8 \\
Lemmatization & Change Acc. & 31.9 & 19.7 & 21.6 & 3.4 & \textbf{36.5} & 11.2 \\
Transliteration & 1-CER & \textbf{92.3} & 92.1 & 88.5 & 20.5 & 91.6 & 90.5 \\
PAWS-Wiki & Accuracy & 62.9 & 55.8 & \textbf{75.2} & 43.0 & 61.4 & 54.6 \\
PIQA & Accuracy & 47.1 & 45.0 & \textbf{50.3} & 16.5 & 14.2 & 46.8 \\
Wiki QA & ROUGE-L & 34.4 & 32.3 & \textbf{34.8} & 16.8 & 34.2 & 27.8 \\
NER (EHRI) & Micro F1 & \textbf{41.3} & 34.2 & 20.8 & 0.8 & 31.7 & 9.2 \\
NER (WikiANN) & Micro F1 & 59.7 & 58.1 & 54.6 & 13.2 & \textbf{62.8} & 40.8 \\
NER (newNLP) & Micro F1 & \textbf{57.6} & 51.9 & 50.5 & 6.4 & 50.9 & 35.8 \\
MT Eng$\to$Yid (FLORES+) & COMET & \textbf{78.5} & 64.8 & 46.5 & 32.9 & 66.7 & 47.2 \\
MT Eng$\to$Yid (\eval-mt) & COMET & \textbf{75.3} & 59.6 & 45.3 & 33.9 & 59.9 & 45.5 \\
MT Yid$\to$Eng (FLORES+) & COMET & \textbf{87.2} & 82.2 & 79.1 & 44.8 & 86.1 & 75.6 \\
MT Yid$\to$Eng (\eval-mt) & COMET & \textbf{79.5} & 72.8 & 72.0 & 44.4 & 77.3 & 70.7 \\
\midrule
\textit{Average} & & \textbf{62.6} & 56.8 & 54.7 & 20.9 & 57.0 & 45.6 \\
\bottomrule
\end{tabular}
\end{table}
 
\section{\eval\ Examples}
\label{app:kashes-examples}
Tables \ref{tab:examples-aya}--\ref{tab:examples-mt} contain examples from each dataset and task in \eval, presented in the format used during evaluation. The quoted English text below each example is provided only as translation and explanation for the reader and is not used as part of the evaluation input.
\begin{table}[h]
\centering
\caption{Examples from the Aya Collection tasks (Originally Machine-translated to Yiddish), formatted as they appear in the few-shot prompts.}
\label{tab:examples-aya}
\small
\begin{tabular}{@{}p{0.95\textwidth}@{}}

\toprule
\textbf{PIQA (Physical Commonsense QA)} \\
\midrule

\textbf{Example 1:} \\
\texttt{Input:} \texthebrew{ענדיג דעם פאלגענדע זאַץ מיט דער בעסטער ברירה:,א מעסער  ברירות: - קענען צעשטערן אַ גומע פּילקע - קענען צעשטערן אַ מאַרמאַרבאָל  ענטפֿערן:} \\
\texttt{Output:} \texthebrew{קענען צעשטערן אַ גומע פּילקע} \\
\textit{``A knife --- can destroy a rubber ball / can destroy a marble.''} \\[4pt]

\textbf{Example 2:} \\
\texttt{Input:} \texthebrew{ענדיג דעם פאלגענדע זאַץ מיט דער בעסטער ברירה: פאַרלענגערן די לעבן פון בלומען אין וואַסע.  ברירות: - לייג אַ קליין סומע פון קאַווע אין וואַזע. - לייג אַ קליין סומע פון 7UP אין וואַזע.  ענטפֿערן:} \\
\texttt{Output:} \texthebrew{לייג אַ קליין סומע פון 7UP אין וואַזע.} \\
\textit{``Extend the life of flowers: add coffee / add 7UP.''} \\

\midrule
\textbf{PAWS-Wiki (Paraphrase Detection)} \\
\midrule

\textbf{Example 1} (\texthebrew{ניין} / No): \\
\texthebrew{זאַץ 1: אַ גרענעץ מלחמה האָט אויסגעבראָכן צווישן מאַגדעבורג און דעם אַרטשבישאָפּ פֿון וואָלפֿענבוטטעל אין 1346. זאַץ 2: אין 1346, איז אויסגעבראָכן אַ גרענעץ־מלחמה צווישן וואָלפֿענבוטל און דעם אַרטשבישאָפּ פֿון מאַגדעבורג. פֿראַגע: צי זענען זאַץ 1 און זאַץ 2 דער זעלביקער טייַטש? יאָ אָדער ניין?} \\
\texthebrew{ניין} \\
\textit{City names swapped between ``X and the Archbishop of Y''---changes who is the city vs.\ the archbishop.} \\[4pt]

\textbf{Example 2} (\texthebrew{ניין} / No): \\
\texthebrew{זאַץ 1: א פערטער קאנדידאט, מיוריעל סמיט, האט זיך צוריקגעצויגן פון די פארמעסט דעם 2טן נאוועמבער, 1979, און געשטיצט ניק טערנעט. זאַץ 2: א פערטער קאנדידאט, ניק טערנעט, האט זיך צוריקגעצויגן פון די פארמעסט דעם 2טן נאוועמבער 1979 און געשטיצט מיוריעל סמיט. פֿראַגע: צי זענען זאַץ 1 און זאַץ 2 דער זעלביקער טייַטש? יאָ אָדער ניין?} \\
\texthebrew{ניין} \\
\textit{``Muriel Smith withdrew and supported Nick Turnet'' vs.\ the reverse---who did what is swapped.} \\

\midrule
\textbf{Wiki QA (Open-ended Question Answering)} \\
\midrule

\textbf{Example 1:} \\
\texthebrew{וואָס איז די פֿראַגע צו: "אַ גוגאָל איז די גרויסע צאָל $10^{100}$; דאָס הייסט, די ציפער 1 נאכגעגאנגען דורך 100 זעראָז:"? די טעמע איז גוגול.} \\
\texthebrew{"וואָס איז אַ גוגל אין מאַטעמאַטיק?"} \\
\textit{Given a passage about googol ($10^{100}$), generate: ``What is a googol in mathematics?''} \\[4pt]

\textbf{Example 2:} \\
\texthebrew{וואָס איז די פֿראַגע צו: "אַ הייס טאָדי, אויך הייס טאָטי, און הייס טאָטי, איז טיפּיקלי אַ געמישט טרינקען געמאכט פון ליקער און וואַסער מיט צוקער און געווירץ און געדינט הייס"? די טעמע איז הייס טאָדי.} \\
\texthebrew{"וואָס איז אין אַ הייס טאָדי?"} \\
\textit{Passage describes a hot toddy recipe; expected question: ``What is in a hot toddy?''} \\

\bottomrule
\end{tabular}
\end{table}

\begin{table}[p]
\centering
\caption{Examples from the UD Yiddish-YiTB treebank tasks, formatted as they appear in the few-shot prompts.}
\label{tab:examples-ud}
\small
\begin{tabular}{@{}p{0.95\textwidth}@{}}

\toprule
\textbf{POS Tagging} \\
\midrule

\textbf{Example 1:} \\
\texttt{Input:} \texthebrew{איך קויף אַ פּאָסטמאַרקע .} \\
\texttt{Output:} \texthebrew{איך}|PRON \texthebrew{קויף}|VERB \texthebrew{אַ}|DET \texthebrew{פּאָסטמאַרקע}|NOUN .|PUNCT \\
\textit{``I buy a postage stamp.''} \\[4pt]

\textbf{Example 2:} \\
\texttt{Input:} \texthebrew{פֿאַרװאָס האָב איך ליב צו לײענען שלום־עליכמען ?} \\
\texttt{Output:} \texthebrew{פֿאַרװאָס}|ADV \texthebrew{האָב}|VERB \texthebrew{איך}|PRON \texthebrew{ליב}|NOUN \texthebrew{צו}|PART \texthebrew{לײענען}|VERB \texthebrew{שלום־עליכמען}|PROPN ?|PUNCT \\
\textit{``Why do I love to read Sholem Aleichem?'' --- V2 word order, light-verb ``hobn lib'', proper noun.} \\

\midrule
\textbf{Dependency Parsing} \\
\midrule

\textbf{Example 1:} \\
\texttt{Input:} {[1]}\texthebrew{איך} {[2]}\texthebrew{קויף} {[3]}\texthebrew{אַ} {[4]}\texthebrew{פּאָסטמאַרקע} {[5]}. \\
\texttt{Output:} {[1]}\texthebrew{איך}(2$\to$nsubj) {[2]}\texthebrew{קויף}(0$\to$root) {[3]}\texthebrew{אַ}(4$\to$det) {[4]}\texthebrew{פּאָסטמאַרקע}(2$\to$obj) {[5]}.(2$\to$punct) \\
\textit{``I buy a postage stamp.'' --- Core relations: nsubj, root, det, obj, punct.} \\[4pt]

\textbf{Example 2:} \\
\texttt{Input:} {[1]}\texthebrew{טאָמס} {[2]}\texthebrew{טאַטע} {[3]}\texthebrew{איז} {[4]}\texthebrew{אַ} {[5]}\texthebrew{ייִד} {[6]}. \\
\texttt{Output:} {[1]}\texthebrew{טאָמס}(2$\to$nmod:poss) {[2]}\texthebrew{טאַטע}(5$\to$nsubj) {[3]}\texthebrew{איז}(5$\to$cop) {[4]}\texthebrew{אַ}(5$\to$det) {[5]}\texthebrew{ייִד}(0$\to$root) {[6]}.(5$\to$punct) \\
\textit{``Tom's father is a Jew.'' --- Copular construction: predicate noun is root, not copula.} \\

\midrule
\textbf{Lemmatization} \\
\midrule

\textbf{Example 1:} \\
\texttt{Input:} \texthebrew{חזירים זענען טאַקע שמוציק ?} \\
\texttt{Output:} \texthebrew{חזירים}$\to$\texthebrew{חזיר} \texthebrew{זענען}$\to$\texthebrew{זײַן} \texthebrew{טאַקע}$\to$\texthebrew{טאַקע} \texthebrew{שמוציק}$\to$\texthebrew{שמוציק} ?$\to$? \\
\textit{``Are pigs really dirty?'' --- Plural ``khazirim''$\to$singular ``khazir''; ``zenen''$\to$infinitive ``zayn''.} \\[4pt]

\textbf{Example 2:} \\
\texttt{Input:} \texthebrew{צי זענט איר אַן איטאַליענערין ?} \\
\texttt{Output:} \texthebrew{צי}$\to$\texthebrew{צי} \texthebrew{זענט}$\to$\texthebrew{זײַן} \texthebrew{איר}$\to$\texthebrew{איר} \texthebrew{אַן}$\to$\texthebrew{אַ} \texthebrew{איטאַליענערין}$\to$\texthebrew{איטאַליענער} ?$\to$? \\
\textit{``Are you an Italian woman?'' --- ``zent''$\to$``zayn'', ``an''$\to$``a'', feminine ``-in'' stripped.} \\

\midrule
\textbf{Transliteration (Hebrew script $\to$ YIVO Latin)} \\
\midrule

\textbf{Example 1:} \\
\texttt{Input:} \texthebrew{דאָס איז זייער אַ שיינער בוים.} \\
\texttt{Output:} dos iz zeyer a sheyner boym. \\
\textit{``This is a very beautiful tree.'' --- ``sh'' for shin, ``ey'' diphthong, ``oy'' for vov-yud.} \\[4pt]

\textbf{Example 2:} \\
\texttt{Input:} \texthebrew{כ'האָב געקויפֿט זײַן נײַסטן ראָמאַן.} \\
\texttt{Output:} kh'hob gekoyft zayn naystn roman. \\
\textit{``I bought his newest novel.'' --- Contracted ``kh'hob'', fey-rafe as ``f'', ``ay'' diphthong.} \\

\bottomrule
\end{tabular}
\end{table}

\begin{table}[t]
\centering
\caption{Examples from the three NER datasets, formatted as they appear in the few-shot prompts.}
\label{tab:examples-ner}
\small
\begin{tabular}{@{}p{0.95\textwidth}@{}}

\toprule
\textbf{EHRI (Holocaust Domain)} --- Entity types: PERSON, LOCATION, ORGANIZATION, CAMP, DATE, GHETTO \\
\midrule

\texttt{Text:} \texthebrew{אין עטליכע טעג ארום איז צו דעם פויער געקומען שמואל מיכל, איטאֵ בראוז און ליובא מעלץ וואס זיינען אויך געווען פריער אַנטלאפן פֿון שול־הויף אין זשאגער.} \\
\texttt{Entities:} \\
\quad PERSON: \texthebrew{שמואל מיכל}; \texthebrew{איטאֵ בראוז}; \texthebrew{ליובא מעלץ} \quad LOCATION: \texthebrew{זשאגער} \\
\textit{``Shmuel Mikhl, Ite Broyz, and Lyuba Melts---who escaped from the synagogue courtyard in Zhager---came to the peasant.'' Three persons and a Lithuanian town from Holocaust testimonies.} \\

\midrule
\textbf{WikiANN} --- Entity types: PERSON, LOCATION, ORGANIZATION \\
\midrule

\texttt{Text:} \texthebrew{יאזעף סטאלין ווערט געוועלטיגער איבערן סאוועטן פארבאנד} \\
\texttt{Entities:} \\
\quad PERSON: \texthebrew{יאזעף סטאלין} \quad LOCATION: \texthebrew{סאוועטן פארבאנד} \\
\textit{``Joseph Stalin becomes ruler over the Soviet Union.'' Wikipedia-sourced, with well-known entities transliterated into Yiddish.} \\

\midrule
\textbf{newNLP NER} --- Entity types: PERSON, LOCATION, ORGANIZATION \\
\midrule

\texttt{Text:} \texthebrew{איבערגעזעצט אויף רוסיש א גאנצע ריי ווערק פון ישראל־יהושע זינגער, יצחק באשעוויס־זינגער, דוד בערגעלסאן, אברהם סוצקעווער און אנדערע יידישע פראזאיקער און דיכטער.} \\
\texttt{Entities:} \\
\quad PERSON: \texthebrew{ישראל־יהושע זינגער}; \texthebrew{יצחק באשעוויס־זינגער}; \texthebrew{דוד בערגעלסאן}; \texthebrew{אברהם סוצקעווער} \\
\textit{``Translated into Russian works by I.J.\ Singer, I.B.\ Singer, Dovid Bergelson, Avrom Sutzkever\ldots'' Four major Yiddish literary figures from modern Yiddish press.} \\

\bottomrule
\end{tabular}
\end{table}

\begin{table}[t]
\centering
\caption{Examples from the machine translation task, formatted as they appear in the few-shot prompts.}
\label{tab:examples-mt}
\small
\begin{tabular}{@{}p{0.95\textwidth}@{}}

\toprule
\textbf{Example 1}: \\
\texttt{English:} The sons, the three thieves, loved their mother very much. \\
\texttt{Yiddish:} \texthebrew{די זין, די דרײַ גנבֿים, האָבן זײער ליב געהאַט די מאַמע.} \\
\textit{Hebrew-origin ``gnavim'' (thieves), idiomatic ``mame'' (mother)---folktale narrative.} \\

\midrule
\textbf{Example 2}: \\
\texttt{English:} ``In that case,'' said the head of the court, ``Let's ask at the prayer house where he used to pray.'' \\
\texttt{Yiddish:} \texthebrew{„אױב אַזױ, — האָט דער ראָש־בית־דין געזאָגט, — זאָל מען שיקן פֿרעגן אין בית־מדרש, װוּ אײַער מאַן פֿלעגט דאַװנען."} \\
\textit{English uses generic terms; Yiddish uses Hebrew-Aramaic loanwords: ``rosh-beys-din'', ``beys-medresh'', ``davnen''.} \\

\bottomrule
\end{tabular}
\end{table}

\newpage
\section{Auditing the Yiddish Split of mC4}
\label{app:mc4-audit}

This appendix details the analysis summarized in \autoref{sec:mc4-analysis}. The split contains 143,708 pages from 7,621 distinct domains, each labelled
\texttt{yi} by the CLD3 identifier used in mC4's construction \citep{xue-etal-2021-mt5}.
We audit all 143,708 pages for two failure modes, Hebrew misidentified as Yiddish through
the shared script, and machine translation; \autoref{tab:mc4-composition-full} gives the
resulting composition.

Hebrew and Yiddish share a script, and general-purpose language identifiers are unreliable
on the pair. We therefore trained a dedicated Hebrew/Yiddish classifier: a character
n-gram Multinomial Naive Bayes model (n = 2--4) trained on 108,000 sentences per class
from Yiddish and Hebrew Wikipedia, with input normalized to Hebrew-script characters and
diacritic-stripped copies added to training. On 12,000 held-out sentences it reaches
macro-F1 0.9968 (95\% CI $[0.9957, 0.9979]$) and stays above 0.995 without diacritics and
on 50-character snippets. 

Because no gold labels exist for the crawl, we validated it against silver labels derived from URL
locale structure (\texttt{yi.}/\texttt{he.}/\texttt{iw.} subdomains and paths), a signal
it does not observe, available for 50,030 pages: it agrees on 99.6\% of them. Applied to
the full split, the classifier marks
31,485 pages (21.9\%) as Hebrew rather than Yiddish (\autoref{tab:mc4-composition-full}).

The second failure mode is machine translation. Machine-translated web content has a
characteristic URL fingerprint: the same site served under many language codes, with the
Yiddish ``edition'' appearing as a \texttt{yi.} or \texttt{yid.} subdomain, a
language-code path segment (\texttt{/yi/}, \texttt{LANG-yi}), or a language-code query
parameter (e.g., \texttt{yi.itsmygame.org}, \texttt{yid.feminineclub.com}). Pages were
assigned to \emph{known Yiddish sources} (manually verified native outlets such as
\texttt{ivelt.com}, \texttt{kaveshtiebel.com}, and \texttt{yiddish.forward.com}) or to
\emph{suspected machine translation} (the URL fingerprint above). We manually inspected
several hundred domains, ranked by page count, and read sampled pages from each category;
the suspected-MT domains were predominantly spam-like sites (gambling, gaming,
explicit-content, and content-farm pages rendered into dozens of languages).

The URL fingerprint alone cannot separate a genuine bilingual outlet from an MT template
site; \texttt{yiddish.forward.com} carries a language-labeled subdomain just as
\texttt{yi.itsmygame.org} does. We therefore backed it with a \emph{locale-sibling count}:
for each domain, we queried a Common Crawl index contemporary with the corpus's
construction and counted the distinct language editions (locale-code subdomains or path
prefixes) under which the same site appears. An MT template site serves the same content
under many language codes, while a native outlet maintains at most one or two editions. On
samples of 50 domains from the suspected-MT and native categories, MT-flagged domains
appeared under a mean of 25 language editions, against fewer than one for native domains.
Genuinely bilingual publishers such as \texttt{yiddish.forward.com} are thus cleared by
measurement rather than by prior knowledge of the Yiddish web.

The locale-sibling count can still produce false positives, since some institutions
legitimately publish in many languages; governmental portals are a typical example, and
some carry genuine Yiddish pages where Yiddish is a recognized minority language. Manual
investigation, however, showed that from the sample almost all domains confirmed by both the fingerprint
and the sibling count came from contexts almost surely unrelated to any Yiddish-speaking
community, such as gambling, gaming, adult-content, and commercial content-farm sites.
While this assessment is somewhat subjective, it is unusually dependable for Yiddish: the
language's present-day native speakers are concentrated in Hasidic and other Haredi
communities, and content of this character is highly unlikely to be produced by or
addressed to that community publicly.

Beyond the pages captured by the native whitelist, the MT fingerprint, and the Hebrew
classifier lies a long tail of 8,871 pages (6.2\% of the split) across some 1,600 domains:
very short Hebrew-script fragments that resist reliable identification, multilingual
template pages, and small uncatalogued native Yiddish sources. Both the native and the MT
shares of \autoref{tab:mc4-composition-full} are therefore lower bounds.

\begin{table}[h]
\centering
\begin{tabular}{lrrr}
\toprule
Category & Pages & \% of split & LK rate \\
\midrule
Known Yiddish sources & 60,587 & 42.2\% & 10.2\% $\pm$ 0.03\% \\
Suspected machine translation & 42,765 & 29.8\% & \phantom{0}3.6\% $\pm$ 0.02\% \\
Hebrew, misidentified as Yiddish & 31,485 & 21.9\% & --- \\
Other & 8,871 & \phantom{0}6.2\% & \phantom{0}9.4\% $\pm$ 0.2\% \\
\bottomrule
\end{tabular}
\caption{Composition of the mC4 Yiddish split, all 143,708 pages. The first two categories
are assigned by source URL; the remaining pages are split by the Hebrew/Yiddish
classifier. LK rate is the mean
per-document share of loshn-koydesh (Hebrew/Aramaic-origin) content words ($\pm$ SE); it
is not reported for the Hebrew category, where Hebrew text produces spuriously high
matches.}
\label{tab:mc4-composition-full}
\end{table}

\noindent
\textbf{More than one page in five in the mC4 Yiddish split is Hebrew, not Yiddish, and
well under half comes from native Yiddish sources.}

Finally, the loshn-koydesh (LK) rate serves as one more independent tool (the lexicon and
matching procedure are described in \hyperref[app:lk-identification]{Appendix~\ref*{app:lk-identification}}). Suspected-MT pages
average an LK rate of 3.6\%, roughly a third of the 10.2\% of native sources, consistent
with translationese that substitutes Germanic or internationalist equivalents for
Hebrew/Aramaic-origin vocabulary. No rate is reported for the Hebrew category, where
Hebrew text trivially matches the Hebrew-origin lexicon. The residual category shows the
mixture its composition implies, with a median LK of 5.7\% against a mean of 9.4\%.

\section{Analysis Methodology}
\label{app:analysis-methodology}

This appendix provides full technical details for the linguistic competence analyses presented in \autoref{sec:analysis}. All analyses draw on two primary resources: a loshn-koydesh (LK) identification lexicon and morphological annotations from the UD Yiddish-YITB treebank.

\subsection{Loshn-Koydesh Identification}
\label{app:lk-identification}

\paragraph{Lexicon.}
We identify Hebrew/Aramaic-origin vocabulary using the Bleaman/Niborski loshn-koydesh pronunciation lexicon,\footnote{\url{https://github.com/ibleaman/loshn-koydesh-pronunciation}} a digitized version of Eliezer Niborski's reference work. The lexicon maps Hebrew orthographic forms to their Yiddish phonetic transcriptions. We extract 5,437 single-word entries and 2,926 compound phrases (entries containing the Hebrew maqaf~\texthebrew{־} or a hyphen delimiter).

\paragraph{Tokenization.}
Yiddish text is tokenized into words using a regular expression over Unicode Hebrew-script code points: characters in the ranges U+05D0--U+05EA (Hebrew letters), U+05F0--U+05F4 (Yiddish ligatures), U+FB1D--U+FB4E (Hebrew presentation forms), and U+05B0--U+05C7 (Hebrew diacritical marks/\textit{nikud}). Each maximal contiguous span of such characters constitutes a token.

\paragraph{Content word filtering.}
Tokens are classified as content words if they satisfy two conditions: (1)~the token contains $\geq$3 characters after removal of all combining diacritical marks (Unicode category Mn), and (2)~the diacritic-stripped form is not in a stopword list of common Germanic-origin function words. The stopword list contains the following items:

\begin{quote}
\small
\texthebrew{%
די, א, אין, צו, פאר, אז, ס, זיין, דאך, צום, פ, מיט, פון, איז, האט, ניט, נישט, אויך, שוין, נאר, ווי, וואס, דאס, דאָס, איך, ער, זי, מיר, איר, דער, דעם, עס, מען, זײ, אים, זיך, אַ, אַן, און, אָדער, אָבער, דאָ, נאָך, פֿון, פֿאַר%
}
\end{quote}

\paragraph{LK matching.}
Each content word is tested against the lexicon using a two-pass procedure:
\begin{enumerate}
    \item \textbf{NFC-normalized exact match}: the token is Unicode NFC-normalized and compared to the set of NFC-normalized lexicon entries.
    \item \textbf{Diacritic-stripped fallback}: if no NFC match is found, all combining marks (Unicode category Mn) are removed from both the token and lexicon entries, and comparison is repeated. This handles variation in \textit{nikud} (vowel pointing) between sources.
\end{enumerate}
For compound phrases, a sliding window of size $n$ (for each $n$-gram size present in the lexicon) is passed over the full token stream. If an $n$-gram matches a compound entry (by NFC or stripped comparison), it is recorded as a single LK item. Individual components of compound entries are \emph{not} treated as standalone LK words, preventing false positives from Germanic words that happen to appear within LK compounds.

\subsection{LK Vocabulary Production in Translation}
\label{app:lk-translation}

This analysis measures the rate at which models produce loshn-koydesh vocabulary in English$\rightarrow$Yiddish translation, corresponding to the LK production results in \autoref{sec:analysis}.

\paragraph{Data.}
We use 5,287 English--Yiddish sentence pairs from the \eval-mt{} dataset, drawn from the In Geveb (4,292) and Forward (995) sources. Each sentence pair is translated by two models (\mllm\  and Llama~3.1~8B) at a 5-shot setup.

\paragraph{Metrics.}
For each generated translation, we apply the tokenization, content word filtering, and LK matching pipeline described in \hyperref[app:lk-identification]{Appendix~\ref*{app:lk-identification}}. We then compute three metrics:

\begin{itemize}
    \item \textbf{LK content word rate}: the proportion of content words identified as LK in the generated text. Computed per sentence and reported as the mean $\pm$ standard error across all $n = 5{,}287$ sentences:
    \[
        \text{LK rate} = \frac{1}{n}\sum_{i=1}^{n} \frac{|\{\text{LK tokens in sentence } i\}|}{|\{\text{content tokens in sentence } i\}|}
    \]

    \item \textbf{LK sentence match rate}: restricted to the $n_{\text{LK}} = 3{,}112$ sentences where the gold reference contains $\geq$1 LK content word. For each such sentence, we check whether the model's translation contains at least one LK word that also appears in the gold (after diacritic-stripped comparison). Reported as:
    \[
        \text{Match rate} = \frac{|\{i : \text{model LK}_i \cap \text{gold LK}_i \neq \emptyset\}|}{n_{\text{LK}}}
    \]

    \item \textbf{Per-word LK recall}: for each gold-LK sentence, the fraction of unique gold LK word types reproduced by the model (after diacritic stripping). Reported as the mean $\pm$ SE across gold-LK sentences:
    \[
        \text{Recall} = \frac{1}{n_{\text{LK}}}\sum_{i=1}^{n_{\text{LK}}} \frac{|\text{model LK}_i \cap \text{gold LK}_i|}{|\text{gold LK}_i|}
    \]
\end{itemize}

\paragraph{LK$\rightarrow$Germanic substitution analysis.}
To characterize what models produce instead of LK words, we examined all 72 LK words with gold frequency $\geq 20$ at 5-shot. For each LK word, we collected all sentences where (a)~the gold contains the LK word but (b)~the model's output does not. From these sentences, we extracted the model's non-gold content words as candidate substitutions. Germanic equivalents were identified \emph{empirically from model output}---not pre-specified---and then manually verified as genuine synonym pairs. 

\paragraph{Statistical tests.}
We compare \mllm\ and Llama~3.1~8B using two paired tests:
\begin{itemize}
    \item \textbf{LK rate}: paired $t$-test on per-sentence LK rates (same sentence, two models).
    \item \textbf{Match rate}: McNemar's test on sentence-level binary outcomes ($b$ = \mllm\ matches but Llama does not; $c$ = Llama matches but \mllm\ does not). Implemented as a two-sided exact binomial test: $p = \texttt{binomtest}(b,\, b+c,\, 0.5)$.
    \item \textbf{Recall}: paired $t$-test on per-sentence recall values.
\end{itemize}

\subsection{Morphological Analysis via Lemmatization}
\label{app:morphology}

This analysis uses lemmatization as a controlled probe of morphological knowledge. The task is to map each inflected Yiddish word to its citation form (lemma), where each token presents a single morphological operation with an unambiguous gold answer. This corresponds to the morphological results in \autoref{sec:analysis}.

\paragraph{Data.}
The test set consists of 929 sentences (7,499 tokens) from the UD Yiddish-YITB treebank.\footnote{\url{https://github.com/UniversalDependencies/UD_Yiddish-YITB}} Gold annotations include surface form, lemma, universal POS tag (UPOS), and dependency relations in CoNLL-U format. Each model generates lemmatization predictions in a structured word$\rightarrow$lemma format at 5-shot counts. 

\paragraph{Prediction postprocessing.}
Model outputs are postprocessed before evaluation: (1)~trailing punctuation characters (\texttt{.,;:!?\"')]\ldots}) are stripped from predicted lemmas using a character-level regex; (2)~indefinite articles (\texthebrew{אַ}/\texthebrew{אַן}) preceding a word are detached and evaluated as separate pairs.

\paragraph{Token classification.}
Every token in the test set is categorized along two axes:
\begin{enumerate}
    \item \textbf{Etymology}: LK vs.\ non-LK. A token is classified as LK if either its surface form or its gold lemma matches the Bleaman/Niborski lexicon (after NFC normalization and diacritic-stripped fallback), and neither form is in the stopword list. This dual check ensures that tokens whose lemma is LK but whose surface form has diverged (or vice versa) are captured.
    \item \textbf{Transformation type}: \emph{change} tokens (surface $\neq$ lemma) vs.\ \emph{identity} tokens (surface $=$ lemma). Change tokens are the informative subset, since identity tokens can be trivially handled by copying the input.
\end{enumerate}

Dataset composition: 565 LK tokens (7.5\%) and 6,934 non-LK tokens (92.5\%). Among LK tokens, 48.1\% require a lemma change vs.\ 33.5\% of non-LK tokens.

\paragraph{LK lemmatization metrics.}
We report \textbf{change accuracy} separately for LK and non-LK strata:
\[
    \text{ChangeAcc}_{\text{stratum}} = \frac{|\{t \in \text{stratum} : t_{\text{word}} \neq t_{\text{lemma}} \wedge t_{\text{pred}} = t_{\text{lemma}}\}|}{|\{t \in \text{stratum} : t_{\text{word}} \neq t_{\text{lemma}}\}|}
\]
The LK--non-LK gap is the arithmetic difference between the two strata's change accuracies.

\subsection{Morphological Category Breakdown}
\label{app:morph-categories}

To understand \emph{which} morphological phenomena drive the overall gap, we classify all 2,593 change tokens into six categories using the gold CoNLL-U UPOS tags for disambiguation:

\begin{enumerate}
    \item \textbf{\textit{ge-} participle} ($n=236$): the surface form begins with \texthebrew{גע} (or a verbal prefix followed by \texthebrew{גע}, e.g., \texthebrew{אָפּגע}, \texthebrew{אױסגע}, \texthebrew{אַרײַנגע}), the lemma does not contain \texthebrew{גע}, and the UPOS is VERB or AUX. We recognize 23 prefixed \textit{ge-} patterns (e.g., \texthebrew{אָפּגע}, \texthebrew{אױסגע}, \texthebrew{אַרײַנגע}, \texthebrew{אַרױסגע}, \texthebrew{אונטערגע}, \texthebrew{צוגע}, among others).

    \item \textbf{Adjective declension} ($n=173$): the surface form ends in \texthebrew{ער}, \texthebrew{ע}, or \texthebrew{ן}; stripping that suffix yields the lemma; and UPOS is ADJ.

    \item \textbf{Hebrew-origin plural} ($n=53$): the surface form ends in \texthebrew{ים} or \texthebrew{ות} (Hebrew masculine/feminine plural suffixes) while the lemma does not, and UPOS is NOUN or ADJ.

    \item \textbf{Determiner paradigm} ($n=291$): the gold lemma is \texthebrew{דער} and the surface form is one of the declined forms \texthebrew{די}, \texthebrew{דאָס}, or \texthebrew{דעם}.

    \item \textbf{Verb conjugation} ($n=1{,}084$): the lemma ends in \texthebrew{ן} or \texthebrew{ען}, UPOS is VERB or AUX, and the token does not fall into the \textit{ge-} participle category. Both surface form and lemma must be $\geq$3 characters.

    \item \textbf{Other} ($n=793$): all remaining change tokens (noun plurals with Germanic suffixes, pronoun inflections, spelling variants, etc.).
\end{enumerate}

Categories are tested in priority order (1--6), so each token is assigned to exactly one category. Change accuracy is computed per category as defined in \hyperref[app:morphology]{Appendix~\ref*{app:morphology}}.

\subsection{Determiner Case System}
\label{app:determiners}

Yiddish has a three-gender, four-case definite article system in which all declined forms share the citation lemma \texthebrew{דער}. We extract all tokens from the lemmatization test set where the gold lemma is \texthebrew{דער}, grouped by surface form:

\begin{center}
\small
\begin{tabular}{llr}
\toprule
Form & Case/Gender & Count \\
\midrule
\texthebrew{די} & Feminine / Plural & 137 \\
\texthebrew{דער} & Masc.\ Nominative (identity) & 104 \\
\texthebrew{דעם} & Masc./Neut.\ Dative/Accusative & 79 \\
\texthebrew{דאָס} & Neuter Nom./Acc. & 75 \\
\bottomrule
\end{tabular}
\end{center}

Lemmatization accuracy is computed per form. For statistical testing, we exclude the identity case (\texthebrew{דער}$\rightarrow$\texthebrew{דער}) and test each declined form individually and aggregated across all three declined forms using McNemar's exact test.

\subsection{Auxiliary Verb Paradigms}
\label{app:auxiliaries}

We extract all tokens tagged as AUX in the gold CoNLL-U annotations (1,000 tokens total) and group them by gold lemma into three paradigms:

\begin{center}
\small
\begin{tabular}{llr}
\toprule
Paradigm & Key forms & Tokens \\
\midrule
\texthebrew{זײַן} (to be) --- suppletive & \texthebrew{איז}, \texthebrew{זענען}, \texthebrew{געווען}, \texthebrew{בין} & 460 \\
\texthebrew{האָבן} (to have) --- regular & \texthebrew{האָט} & 163 \\
\texthebrew{װעלן} (will) & \texthebrew{װעט} & 50 \\
\bottomrule
\end{tabular}
\end{center}

The key comparison is between \texthebrew{זײַן} (suppletive: inflected forms bear no surface resemblance to the lemma) and \texthebrew{האָבן} (regular: \texthebrew{האָט}$\rightarrow$\texthebrew{האָבן} is a straightforward suffix change). This serves as a within-task control: if a model's advantage is specific to language-specific knowledge rather than general lemmatization ability, it should appear on the suppletive paradigm but not the regular one.

\subsection{Statistical Testing}
\label{app:statistical-tests}

All pairwise model comparisons (\mllm\ vs.\ Llama~3.1~8B) on lemmatization use \textbf{McNemar's exact test} (two-sided). The test operates on paired binary outcomes: for each token (or sentence, in the translation analysis), we record whether each model's prediction is correct. We count:
\begin{itemize}
    \item $b$: tokens where \mllm\ is correct and Llama is incorrect (discordant, favoring \mllm)
    \item $c$: tokens where Llama is correct and \mllm\  is incorrect (discordant, favoring Llama)
\end{itemize}
Concordant pairs (both correct or both incorrect) are uninformative and excluded. Under the null hypothesis that both models are equally likely to be correct on discordant pairs, $b \sim \text{Binomial}(b+c,\, 0.5)$. The two-sided $p$-value is computed via \texttt{scipy.stats.binomtest(b, b+c, 0.5)}. We report significance at three levels: ${*}$~($p < 0.05$), ${**}$~($p < 0.01$), ${***}$~($p < 0.001$).

McNemar's test is appropriate here because observations are paired (both models predict on the same token) and outcomes are binary (correct/incorrect). It is more powerful than unpaired tests because it controls for item difficulty.

For the translation analysis, we additionally use \textbf{paired $t$-tests} on continuous per-sentence metrics (LK rate, LK recall), since these are real-valued rather than binary.

\subsection{Summary of All Metrics}
\label{app:full-summary}

\autoref{tab:full-summary} consolidates all metrics from the analyses above, comparing \mllm{} and Llama~3.1~8B across both the LK translation analysis (5-shot) and the morphological analysis (5-shot lemmatization).

\begin{table*}[t]
\centering
\caption{Comparison of \mllm{} and Llama~3.1~8B across all analysis
dimensions. Translation metrics are at 5-shot ($n=5{,}287$ sentence
pairs; match and recall use $n=3{,}112$ gold-LK sentences).
Morphological metrics are at 5-shot on UD Yiddish-YITB.
McNemar $b$:$c$ denotes discordant pairs favoring \mllm{} versus Llama~3.1.
All $p$-values are two-sided.}
\label{tab:full-summary}

\small
\setlength{\tabcolsep}{4pt}
\begin{tabularx}{\textwidth}{
    @{}X
    r
    r
    >{\raggedleft\arraybackslash}p{2.0cm}
    >{\raggedleft\arraybackslash}p{1.8cm}
    @{}
}
\toprule
\textbf{Metric}
& \textbf{\mllm}
& \textbf{Llama 3.1}
& \makecell[r]{\textbf{Test}\\\textbf{statistic}}
& \textbf{$p$-value} \\
\midrule

\multicolumn{5}{@{}l}{%
  \textit{LK vocabulary production (English$\to$Yiddish translation)}} \\
LK content-word rate (\%; gold: 6.2)
  & 4.7 & 1.6 & $t=28.0$ & $<10^{-161}$ \\
LK sentence-match rate (\%)
  & 52.4 & 16.0 & $b$:$c=1181{:}47$ & $<10^{-229}$ \\
Per-word LK recall (\%)
  & 35.6 & 9.3 & $t=36.8$ & $<10^{-246}$ \\

\midrule
\multicolumn{5}{@{}l}{%
  \textit{Morphological analysis (lemmatization)}} \\
\textit{ge-} participles ($n=236$)
  & 50.8 & 5.1 & $110{:}2$ & $<10^{-29}$ \\
Hebrew-origin plurals ($n=53$)
  & 30.2 & 2.3 & $12{:}0$ & $<0.001$ \\
Adjective declension ($n=173$)
  & 68.5 & 41.8 & $42{:}3$ & $<10^{-9}$ \\
Determiners ($n=291$)
  & 28.2 & 15.5 & $55{:}18$ & $<10^{-5}$ \\
Verb conjugation ($n=1{,}084$)
  & 46.7 & 33.8 & $196{:}56$ & $<10^{-18}$ \\

\midrule
\multicolumn{5}{@{}l}{%
  \textit{Auxiliary verb paradigms (lemmatization)}} \\
\texthebrew{זײַן} --- suppletive ($n=460$)
  & 20.0 & 8.9 & $63{:}12$ & $<10^{-9}$ \\
\texthebrew{האָבן} --- regular ($n=163$)
  & 84.7 & 84.7 & $14{:}14$ & $1.0$ (n.s.) \\

\bottomrule
\end{tabularx}
\end{table*}

\subsection{Most Frequent Loshn-Koydesh Words in Gold References}
\label{app:lk-top-words}

\autoref{tab:lk-top15} lists the 15 most frequent LK content words in the gold Yiddish references, along with the absolute number of times each word appears in the 5-shot translations of \mllm{} and Llama~3.1~8B (across all $n{=}5{,}287$ sentence pairs from the In Geveb and Forward sources).

\begin{table}[h]
\centering
\caption{Top 15 LK words by gold frequency (absolute counts across 5,287 5-shot translations). Common Yiddish vocabulary items (e.g., \textit{efsher}, \textit{ponim}, \textit{kedey}) are largely absent from Llama's output, while proper nouns and cultural terms (e.g., \textit{rebe}, \textit{khane}, \textit{yisroel}) appear at comparable rates.}
\label{tab:lk-top15}
\small
\begin{tabular}{llllrrr}
\toprule
\textbf{Word} & \textbf{Translit.} & \textbf{Meaning} & \textbf{Type} & \textbf{Gold} & \textbf{\mllm} & \textbf{Llama 3.1} \\
\midrule
\texthebrew{בין} & \textit{bin} & between & Common & 259 & 241 & 84 \\
\texthebrew{אפילו} & \textit{afile} & even & Common & 153 & 82 & 19 \\
\texthebrew{אפשר} & \textit{efsher} & perhaps & Common & 99 & 61 & 0 \\
\texthebrew{אסתר} & \textit{ester} & Esther & Proper noun & 96 & 16 & 21 \\
\texthebrew{עולם} & \textit{oylem} & world & Common & 85 & 42 & 10 \\
\texthebrew{פנים} & \textit{ponim} & face & Common & 83 & 65 & 0 \\
\texthebrew{רבי} & \textit{rebe} & Rabbi & Cultural & 67 & 67 & 73 \\
\texthebrew{כדי} & \textit{kedey} & in order to & Common & 66 & 61 & 0 \\
\texthebrew{אמת} & \textit{emes} & truth & Common & 64 & 54 & 2 \\
\texthebrew{חנה} & \textit{khane} & Chana & Proper noun & 62 & 49 & 50 \\
\texthebrew{חתונה} & \textit{khasene} & wedding & Common & 57 & 46 & 10 \\
\texthebrew{נשמה} & \textit{neshome} & soul & Common & 56 & 42 & 14 \\
\texthebrew{ישראל} & \textit{yisroel} & Israel & Proper noun & 55 & 44 & 50 \\
\texthebrew{שעה} & \textit{sho} & hour & Common & 48 & 32 & 5 \\
\texthebrew{שבת} & \textit{shabes} & Shabbat & Cultural & 47 & 32 & 36 \\
\bottomrule
\end{tabular}
\end{table}

\section{Effect of related-language mixing}
\label{app:mixtures}
\begin{table}[H]
\centering
\caption{Results over \eval\ of \mllm\ and Llama 3.1 compared to models trained on data mixtures with more Yiddish-related languages. All results are in a 5-shot setting, scaled 0--100. \textbf{Bold} = best per row.}
\label{tab:multiling-exp}
\small
\setlength{\tabcolsep}{4pt}
\begin{tabular}{llcccc}
\toprule
Task & Metric & \mllm & Llama 3.1 & Rebalanced & Expanded \\
\midrule
POS Tagging & Accuracy & \textbf{88.6} & 86.9 & 61.6 & 66.4 \\
Dep.\ Parsing & LAS & \textbf{40.6} & 39.7 & 15.6 & 25.3 \\
Lemmatization & Change Acc. & 31.9 & 19.7 & \textbf{33.2} & 23.2 \\
Transliteration & 1--CER & 92.3 & 92.1 & 93.4 & \textbf{93.4} \\
PAWS-Wiki & Accuracy & \textbf{62.9} & 55.8 & 53.2 & 55.9 \\
PIQA & Accuracy & \textbf{47.1} & 45.0 & 46.9 & 46.0 \\
Wiki QA & ROUGE-L & \textbf{34.4} & 32.3 & 34.0 & 34.4 \\
NER (EHRI) & Micro F1 & \textbf{41.3} & 34.2 & 25.3 & 30.1 \\
NER (WikiANN) & Micro F1 & \textbf{59.7} & 58.1 & 55.5 & 58.6 \\
NER (newNLP) & Micro F1 & \textbf{57.6} & 51.9 & 34.5 & 41.8 \\
MT Eng$\to$Yid (FLORES+) & COMET & 78.5 & 64.8 & \textbf{79.3} & 79.2 \\
MT Eng$\to$Yid (\eval-mt) & COMET & 75.3 & 59.6 & \textbf{76.0} & 75.9 \\
MT Yid$\to$Eng (FLORES+) & COMET & 87.2 & 82.2 & 87.1 & \textbf{87.3} \\
MT Yid$\to$Eng (\eval-mt) & COMET & \textbf{79.5} & 72.8 & 79.0 & 79.0 \\
\midrule
\textit{Average} & & \textbf{62.6} & 56.8 & 55.3 & 56.9 \\
\bottomrule
\end{tabular}
\end{table}

\begin{table}[H]
\centering
\caption{Training mixture composition across the three continued-pretraining settings. Percentages are shown over words and over training tokens.}
\label{tab:mix-summary}
\small
\setlength{\tabcolsep}{4pt}
\begin{tabular}{lcccccc}
\toprule
& \multicolumn{2}{c}{\mllm} & \multicolumn{2}{c}{Rebalanced} & \multicolumn{2}{c}{Expanded} \\
\cmidrule(lr){2-3} \cmidrule(lr){4-5} \cmidrule(lr){6-7}
Language & Words (\%) & Tokens (\%) & Words (\%) & Tokens (\%) & Words (\%) & Tokens (\%) \\
\midrule
Yiddish  & 72.0 & 91.8 & 72.7 & 85.2 & 48.8 & 71.0 \\
English  & 28.0 & 8.2  & 14.2 & 3.8  & 19.0 & 6.4  \\
Hebrew   & --   & --   & 7.1  & 8.3  & 4.8  & 6.9  \\
German   & --   & --   & 6.0  & 2.4  & 10.0 & 5.0  \\
Russian  & --   & --   & --   & --   & 9.7  & 5.6  \\
Polish   & --   & --   & --   & --   & 7.8  & 5.3  \\
\midrule
Total    & 100  & 100  & 100  & 100  & 100  & 100  \\
\bottomrule
\end{tabular}
\end{table}

\end{document}